\documentclass[10pt,twocolumn,letterpaper]{article}

\usepackage{cvpr}
\usepackage{times}
\usepackage{epsfig}
\usepackage{graphicx}
\usepackage{amsmath}
\usepackage{amssymb}

\usepackage{booktabs}
\usepackage{multirow}
\usepackage{adjustbox}

\newcommand{\ignore}[1]{}
\makeatletter
\def\ie{\emph{i.e.}}

\def\etal{\emph{et al.}}
\makeatother

\usepackage[pagebackref=true,breaklinks=true,letterpaper=true,colorlinks,bookmarks=false]{hyperref}

\cvprfinalcopy % *** Uncomment this line for the final submission

\def\cvprPaperID{562} % *** Enter the CVPR Paper ID here

\ifcvprfinal\fi
\begin{document}

%%%%%%%%% TITLE
\title{Super-Resolution with Deep Adaptive Image Resampling}
% Single Image Super-Resolution with Deep Adaptive Image Resampling
% Revisiting resampling method for single image super-resolution with deep learning

\author{
  Xu Jia$^{1}$\\
  KU Leuven ESAT-PSI/IMEC\\   
  \and
  Hong Chang$^{2}$\\
  Key Lab of Intelligent Information Processing,\\
  Institute of Computing Technology, \\
  Chinese Academy of Sciences (CAS)\\
  \and
  Tinne Tuytelaars$^{1}$\\
  KU Leuven ESAT-PSI/IMEC\\
  $^{1}${\tt\small{firstname.lastname@esat.kuleuven.be}} \quad $^{2}${\tt\small{hong.chang@vipl.ict.ac.cn}}\\
}

\maketitle
%\thispagestyle{empty}

%%%%%%%%% ABSTRACT
\begin{abstract}
   Deep learning based methods have recently pushed the state-of-the-art on the problem of Single Image Super-Resolution (SISR). In this work, we revisit the more traditional interpolation-based methods, that were popular before, now with the help of deep learning. In particular, we propose to use a Convolutional Neural Network (CNN) to estimate spatially variant interpolation kernels and apply the estimated kernels adaptively to each position in the image. % a low resolution image. 
The whole model is trained in an end-to-end manner. We explore two ways to improve the results for the case of large upscaling factors, and propose a recursive extension of our basic model. This achieves results that are on par with state-of-the-art methods.
% the state-of-the-art.
   %Several ways are explored to further improve the performance. 
   %We evaluate the proposed method on benchmark datasets and show that it performs comparable to state-of-the-art methods. 
We visualize the estimated adaptive interpolation kernels to gain more insight on the effectiveness of the proposed method. We also extend the method to the task of joint image filtering and again achieve state-of-the-art performance.
\end{abstract}

%%%%%%%%% BODY TEXT
%%%%%%%%% BODY TEXT
\section{Introduction}
\label{sec:intro}
% sampling based methods (bilinear, bicubic, lanczos)
% sparse coding/dictionary based methods
% deep learning based methods 
% (directly conv to generate super-res result from scratch like srcnn or espcnn;
% use cnn to compute residual and combined with precomputed bicubic interpolation result as result)

Reconstructing a high-resolution (HR) image from a low-resolution (LR) input is a classic computer vision problem, referred as {\em Single Image Super-Resolution} (SISR). Research on SISR receives a lot of attention because of the wide range of applications, such as surveillance, medical imaging and remote sensing imaging, where high-frequency details are required. 
% * <changhong@ict.ac.cn> 2017-11-13T21:10:53.494Z:
%
% ^.
The main difficulty with SISR lies in the fact that it is an ill-posed problem: the high-frequency information is missing and there are many possible solutions that are all consistent with the given low-resolution input. Therefore, additional assumptions have to be made regarding the formation of HR images. A common key assumption for this task is that the high-frequency information is redundant and can be reconstructed either from the given LR image or from external exemplars.

% theory motivated methods, interpolation/filtering
Long-standing, basic methods for SISR are general {\em interpolation based methods}, such as bilinear interpolation, bicubic interpolation and Lanczos resampling~\cite{Blu-tip04}. These methods are motivated either by the sampling theorem or spline theory. 
%In Nyquist-Shannon sampling theory, the optimal reconstruction filter for band-limited continuous signals is the sinc filter and the reconstruction is a linear combination of sinc functions. 
%However, the sinc interpolation kernel is impossible to implement since it has infinite support. Therefore, 
%In practice, Lanczos resampling is used as a finite-length approximation of the infinite sinc function. 
%(\hong{The above description is not very relevant so could be more concise.} \xu{This is brief intro of motivation of interpolation based methods and what is an ideal interpolation in theory.})
% TT: I would leave it in, unless we need more space
While they have a strong theoretical basis, they %these traditional interpolation methods 
assume a band-limited continuous signal and apply a fixed interpolation kernel to the LR image to achieve the upscaling. As a result, they cannot adapt to the image content, often resulting in aliasing artefacts or over-smoothed regions. 
To address this issue, several works~\cite{Zhang-tip06,Takeda-tip07,Zhou-ietip12} have proposed {\em edge guided image interpolation methods}. They use prior information about the images as regularization such that they can upscale the image while keeping the edges sharp. 
% In a similar spirit, He~\etal~\cite{He-tpami13} proposed a guided image filter where spatially variant filters are computed for an edge-preserving smoothing operation. 
% This is applied to several image editing tasks, including SISR. (it is not applied to SISR but to depth image super-resolution or colorization image super-resolution)

% learning based / data-driven based methods
More recently, learning based, i.e. data-driven, methods have become more popular.
%Recent methods which work well on SISR learn a mapping between LR space and HR space in a data-driven fashion. 
%One family of methods falling in this category are 
This includes {\em dictionary based methods}~\cite{Chang-cvpr04, Yang-cvpr08, Timofte-iccv13, Timofte-accv14,Yang-eccv14-survey}. They explicitly learn a dictionary mapping between LR space and HR space. Once the mapping is learned, the same set of coding coefficients computed for the LR image are used for the HR image to produce the super-resolved result. 
Another family of data-driven methods are {\em deep learning based models}~\cite{Dong-tpami16, Dong-eccv16, Shi-cvpr16}. Building on the powerful capability of deep neural networks to approximate arbitrary functions, these methods learn an implicit mapping between LR and HR images, typically with a fully-convolutional network and in an end-to-end manner. Deeper networks~\cite{Kim-cvpr16-vdsr, Kim-cvpr16-drcn, Mao-nips16,Caballero-cvpr17} have been proposed to further improve the performance and currently define the state-of-the-art. 
% (\hong{One or two survey papers can be cited here: C.-Y. Yang, C. Ma, and M.-H. Yang. Single-image superresolution: a benchmark. In ECCV, 2014. K. Nasrollahi and T. B. Moeslund. Super-resolution: A comprehensive survey. Machine Vision and Applications, 25(6):1423–1468, 2014.} \xu{I add Ming-Hsuan's paper when I describe dictionary based methods because both of these do not include DL model})

\begin{figure*}[t]
    \begin{tabular}{cc}
    	\begin{adjustbox}{valign=c}
			\begin{tabular}{c}
				\includegraphics[width=0.45\linewidth]{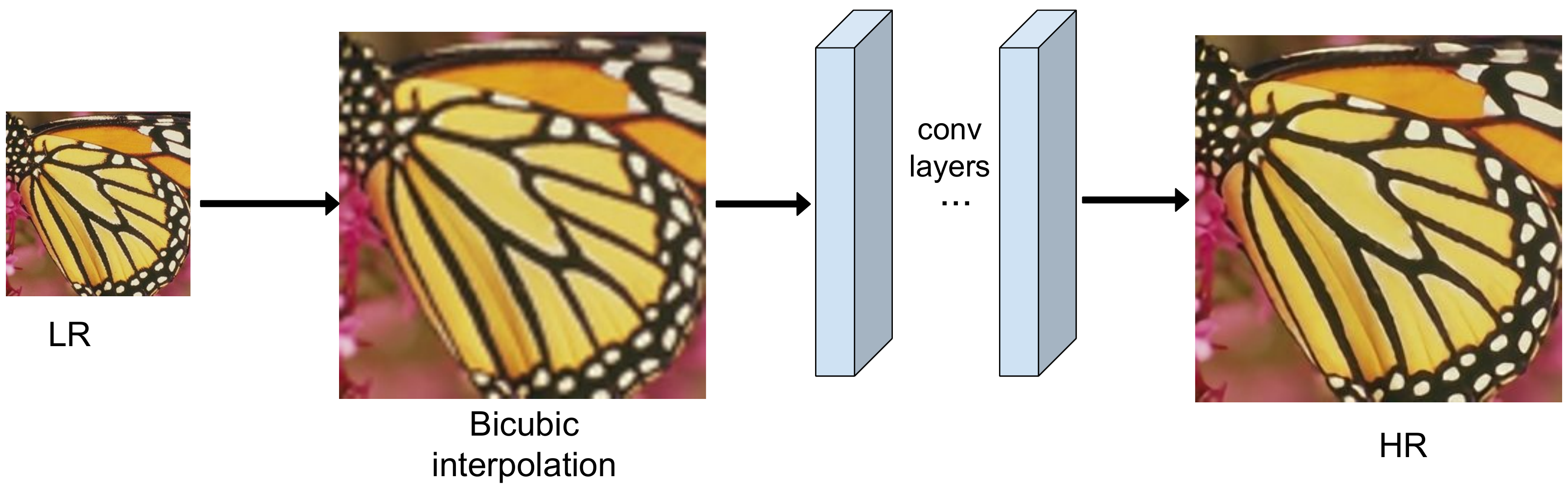}
				\\[1ex]
				(a) SRCNN~\cite{Dong-tpami16}
				\\[1ex]
				\includegraphics[width=0.45\linewidth]{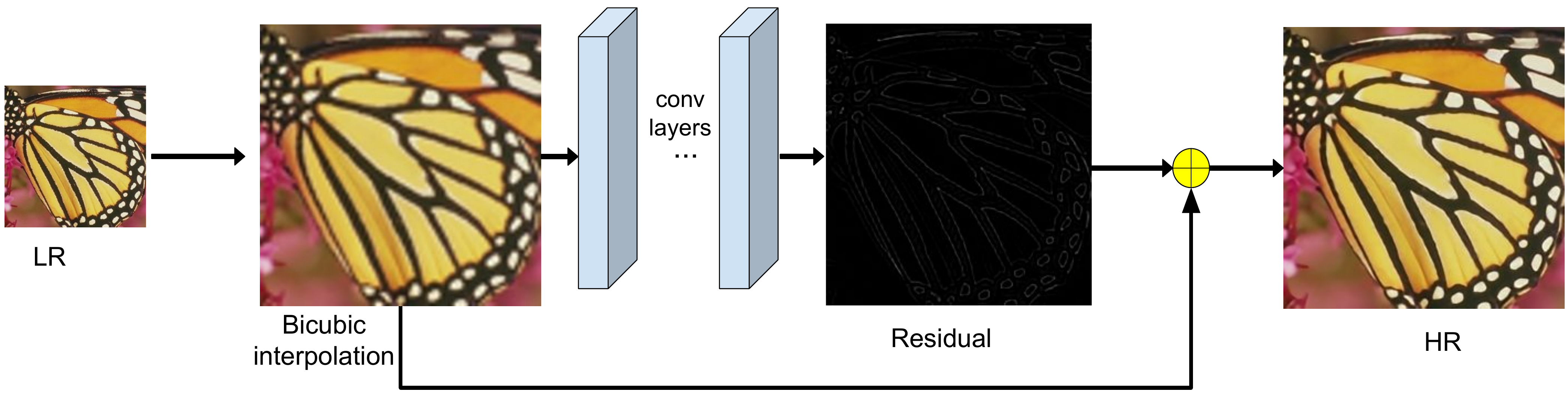}
				\\[1ex]
				(b) VDSR~\cite{Kim-cvpr16-vdsr}
				\end{tabular}
		\end{adjustbox}
		&
        \begin{adjustbox}{valign=c}
			\begin{tabular}{c}
				\includegraphics[width=0.45\textwidth]{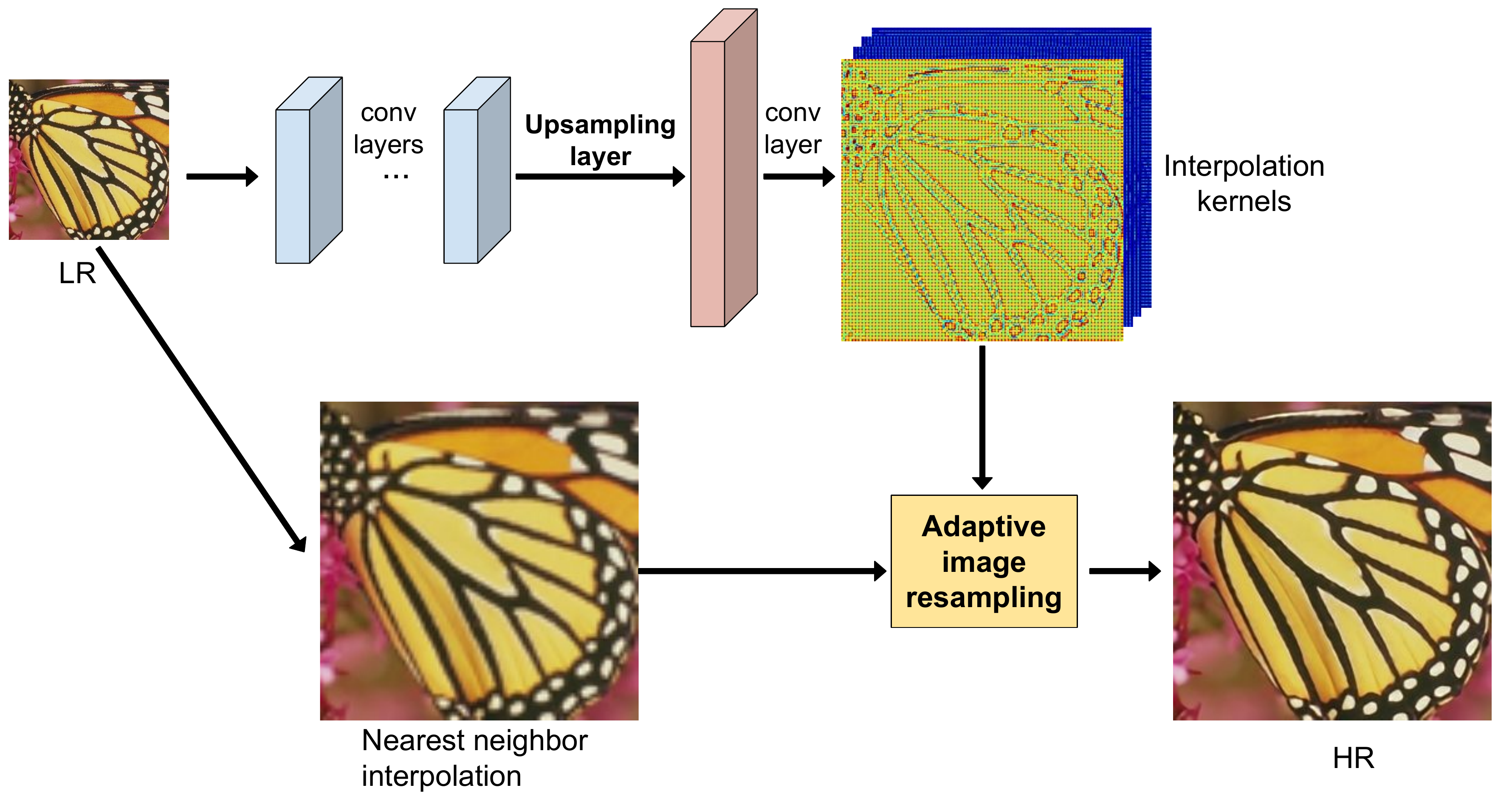}
				\\[1ex]
				(c) Ours
			\end{tabular}
		\end{adjustbox}
    \end{tabular}	
    \caption{Network architecture comparison between our method, SRCNN and VDSR.}
    \label{fig:pipeline}
    \vspace{-3mm}
\end{figure*} 
% Though the above deep learning based models obtain very good super-resolution result, it is not that obvious to know what the NN has learned.
In this work, instead of further increasing the network depth, we revisit the idea of the interpolation based methods, but now with the help of deep learning, aiming at an effective model with some insights. 
% Similar in spirit to the edge guided image interpolation work~\cite{He-tpami13}, (\xu{He-tpami13 is not used for SISR but for joint image filtering})
We compute a pixel in the HR image using adaptive interpolation, \ie a weighted average of the nearby pixels in the corresponding LR image, with weights that are not fixed but depend on the image content at that position. Therefore, the interpolation kernels are spatially variant and content-aware. For example, in smooth regions, there is not much variance among pixels in a neighborhood, so a uniform kernel might do a reasonable job; however, 
% for regions corresponding to edges, kernels dominated by only one or a few non-zero elements, might produce sharper edges; 
for a region with an edge or some rich texture, a specially-designed combination of neighboring pixels is required for its interpolation. 

Instead of using hand-designed kernels to do the filtering/interpolation, we propose to use a deep neural network to learn good interpolation kernels in a data-driven fashion. For this, we build on the recently proposed Dynamic Filter Network architecture~\cite{DeBrabandere-nips16, Finn-nips16, Xue-nips16}.
%
% Deep neural networks are known for their powerful modeling capability. They are able to learn a more advanced mapping function for input and output pairs than methods like the guided image filter, which uses a local linear model~\cite{He-tpami13}. 
% (\hong{This sentence can be moved to the later sections.} \xu{already comment the above sentence})
% For example, deep neural networks are able to recognize not only the pattern of a region (smooth region/edge/textured region) but also its semantic meaning, which may help in figuring out a suitable way to combine its nearby pixels for its super-resolving. 
%
Once the adaptive interpolation kernels are estimated, we use them in an adaptive image resampling layer
which carries out the actual filtering operation
(see Figure~\ref{fig:pipeline} (c)): the estimated interpolation kernels are applied to a (Nearest Neighbour interpolated) low resolution image to obtain the super-resolved result. The adaptive image resampling module is differentiable and allows end-to-end training for the whole model. 
% \hong{It's better to change the order of subfigures in Figure 1.}

% The proposed image resampling layer is similar to the bilinear upsampling layer proposed in~\cite{Jaderberg-nips15}. The added module is differentiable and allows end-to-end training for the whole model. However, in our case the weights are computed by the network and the size of the kernel can be designated as a hyperparameter. 

The performance of interpolation-based methods drops as the upscaling factor increases. That is because when the upscaling factor is big, there is little correlation among nearby pixels. In this case non-local methods perform better than local linear filtering methods. We explore two ways to 
reduce % address 
the degeneration in case of large upscaling factor for interpolation-based methods: an atrous spatial pyramid and progressive upsampling. Besides, the deep adaptive image resampling can be applied to the previously obtained super-resolution result several times, i.e. in a recurrent fashion, to further improve the performance.

The proposed methods are evaluated on four super-resolution benchmark datasets and perform favorably compared to state-of-the-art methods.
% in both accuracy, speed and memory.
We visualize the estimated interpolation kernels and shed some light on why the proposed method works well.
In addition, we show that the proposed method can be naturally extended for the joint image filtering task and again obtains 
% state-of-the-art 
very good performance.

% Contributions:
% a new way to address image super-resolution task with deep learning model;
% analyze what is learned with the deep learning model;
% joint image filtering;

%-------------------------------------------------------------------------
\section{Related Work}
\label{sec:related}
% DL applied to SR task, residual based super-resolution (which do not generate each pixel from scratch), such as VDSR, Deep laplacian pyramid, EDSR;
\paragraph{Deep Learning for Super-Resolution}
Recently, a lot of works have addressed the task of SISR based on Convolutional Neural Networks (CNN). One pioneering work is the Super-Resolution Convolutional Neural Network (SRCNN, see Figure~\ref{fig:pipeline} (a))~\cite{Dong-tpami16,Dong-eccv16}. It implicitly learns a mapping between LR and HR images using a fully-convolutional network. It takes bicubic interpolation as a pre-processing step and feeds the interpolated result to the network for super-resolution. This slows down the processing speed and increases the memory requirement as all convolution operations are done on HR images. The Efficient Sub-Pixel Convolutional Neural Network (ESPCN)~\cite{Shi-cvpr16} addresses this issue by feeding the small size LR image to the network and postponing the upscaling to just before the output layer, by means of a newly proposed sub-pixel layer. Inspired by the success of very deep networks in recognition tasks~\cite{Krizhevsky-nips12-alexnet,Simonyan-iclr15-vggnet,Szegedy-cvpr15-googlenet}, Kim~\etal~\cite{Kim-cvpr16-vdsr,Kim-cvpr16-drcn} proposed Very Deep Super Resolution (VDSR, see Figure~\ref{fig:pipeline} (b)), increasing the network depth to 20 layers. Moreover, 
inspired by ResNet~\cite{He-cvpr16-resnet}, 
they predict the residual between the bicubic interpolation result and the HR image instead of directly predicting the HR image, which eases the training process. Both steps improve performance further.
% Similarly, our method also does not directly estimate the HR image. Instead we estimate the interpolation kernels and then perform the interpolation by resampling nearby pixels.
In~\cite{Mao-nips16}, skip connections are added to the convolutional and deconvolutional layers of very deep convolutional encoder-decoder networks for faster convergence and more detailed restoration.
Lai~\etal~\cite{Lai-cvpr17} proposed the Deep Laplacian Pyramid Network to do the upscaling progressively from small upscaling factor to large upscaling factor.
Very recently, SRResNet~\cite{Caballero-cvpr17}, EDSR~\cite{Lim-cvprws17} and DRRN~\cite{Tai-cvpr2017} proposed to not only use the residual connection in the last layer but also local residual connections in the intermediate layers as in the ResNet~\cite{He-cvpr16-resnet} and DenseNet~\cite{Huang-cvpr17-densenet} architectures, to further improve the performance. A comparison between our network architecture and two popular ones is shown in Figure~\ref{fig:pipeline}. Ours is most similar to the VDSR architecture, 
% except that we do not compute residuals but rather interpolation kernels.
except that we do interpolation instead of addition operation.

% Adaptive convolution, such as dfn based video prediction, adaptive convolution for video interpolation, adaptive convolution for classification, deformable cnn for detection/segmentation, conditioned cnn for super-resolution;
\vspace{-3mm}
\paragraph{Adaptive Convolution}
Very recently several works have proposed to modify the traditional convolutional layer and make it more adaptive to the input, as we do. 
% In the context of SISR, Riegler~\etal~\cite{Riegler-iccv15} proposed to condition the parameters of the first convolutional layer of SRCNN on the input image such that the model can use different blur kernels for different images. 
In the context of image classification, Jeon and Kim~\cite{Jeon-cvpr17} introduced an active convolution unit, which allows a convolutional layer to have flexible shape.
In~\cite{Dai-iccv17}, convolutional layers are further modified such that each position has an adaptive receptive field. This gives good performance on both object detection and semantic segmentation tasks.
Recently,~\cite{Finn-nips16,DeBrabandere-nips16,Xue-nips16} simultaneously proposed the Dynamic Filter Network to model spatial transformations with a single convolution step to address the task of video prediction,
with the model conditioned not only on different input but also on different positions in the image.
Niklaus~\etal~\cite{Niklaus-iccv17} extended this work to the video interpolation task by replacing a single 2D convolution step with two separable 1D convolutions. 
Our work is one of the few pioneering works to relate the idea of adaptive convolution with the SISR task. 
% The closest work 
One similar work
in this context is by Riegler~\etal~\cite{Riegler-iccv15}. 
% However, their purpose and their way of using adaptive convolution are very different: they condition the parameters of the first convolutional layer of SRCNN on the input image in order to address different blur kernels for different images. 
However, they modified SRCNN by conditioning the parameters of its first convolutional layer on the input image in order to address different blur kernels for different images. This requires a different setup, so cannot directly be compared against.
% \hong{the last related work not clear!}. 

\section{Deep Adaptive Image Resampling} % Method
\label{sec:method}
In this section, we describe our proposed deep adaptive image resampling model (section~\ref{sec:basic}) and several further refinements thereof (section~\ref{sec:refinement}). In addition, it is also extended for the joint image filtering task (section~\ref{sec:joint}).

\subsection{The Basic Model}
\label{sec:basic}
Our model is composed of two parts: one module to estimate the adaptive image interpolation kernels, and another module applying the interpolation kernels to the LR input to produce the super-resolved result. The full architecture is shown in Figure~\ref{fig:pipeline} (c).\\

\noindent{\textbf{Adaptive interpolation kernels.}}
Instead of using a fixed blind interpolation kernel for every image and every position, we propose to use a data-driven method to compute a 
% sample-dependent and spatially variant 
content-aware
interpolation kernel separately for each position in the image. We use a fully convolutional network (FCN)~\cite{Long-cvpr15} to compute the weights of the interpolation kernels. 
% TT: left out to gain space
%FCN has been widely used to directly predict a label (semantic segmentation~\cite{Long-cvpr15}) or a value (super-resolution~\cite{Dong-tpami16}) for each position in an image. 
% similar to~\cite{DeBrabandere-nips16}, 
%Here we use FCN to estimate the weights of the interpolation kernels. 
Our FCN consists of several standard convolutional layers and an upsampling layer. 
% We found consecutive standard convolutional layers of $3\times3$ with Relu activation function, but without max-pooling or striding to be effective. 
The convolutional layers in the FCN learn to model local context for each position in an LR image.
% without loss in spatial resolution. 
Its output is a set of feature maps denoted as $K_L$, where $K_L \in \mathbb{R}^{h\times w\times {(f\times s)^2}}$, $h$ and $w$ are the size of the LR input, $f$ is the spatial size of the interpolation kernels
%, as explained later, 
and $s$ is the upscaling factor. $K_L$ has the same spatial resolution as the LR input. To adapt its spatial resolution to HR images, we add an upsampling layer, which can be implemented as either a subpixel layer~\cite{Shi-cvpr16} or a fractionally-strided convolutional layer~\cite{Long-cvpr15}. 
% Though Odena~\etal~\cite{Odena-distill16} reported that several upsampling layers have the risk of producing checkerboard artefacts, in our case, we do not suffer from this issue since we do not directly use the upsampling layers to produce the final image but rather to estimate intermediate interpolation kernels. 

\begin{figure}[t]
	\includegraphics[width=0.95\linewidth]{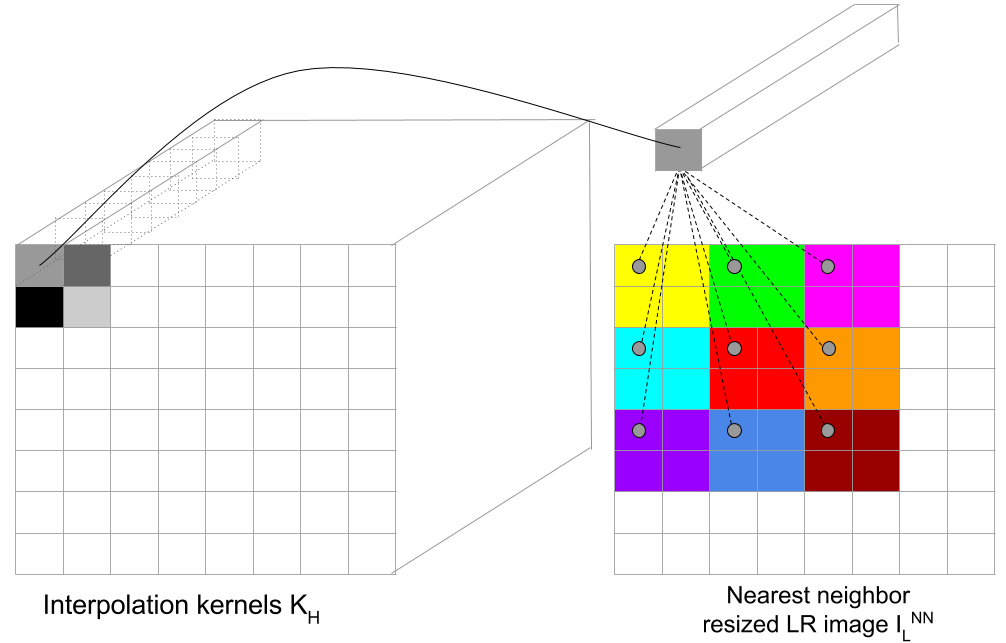}
    \caption{Demonstration of the adaptive image resampling layer for upscaling by a factor 2 and filter size 3x3.}
    %\hong{``dynamic filter'' in the figure}\xu{Sorry, my mistake. Now I upload a new figure.}
    %\hong{better to change the ``dynamic filtering layer''}\xu{thank you for pointing such mistake!}
    \label{fig:upscaling}
%     \vspace{-5mm}
\end{figure}

The upsampled interpolation kernels are denoted as $K_H$, where $K_H \in \mathbb{R}^{(h\times s)\times (w\times s)\times {f^2}}$
and $K_H = FCN(I_L)$. 
Each spatial position in $K_H$ corresponds to a vector of dimension $f^2$. It can be reshaped to a filter of size $f\times f$ and works as an interpolation kernel at that position. The interpolation kernel combines the nearby $f^2$ pixels in the LR input and reconstructs the corresponding pixel in the HR image. The interpolation estimation module is expected to learn which elements in a neighborhood contribute to the reconstruction of a certain pixel and how much each of them contributes.\\
% (\hong{better to indicate $K_L$ and $K_H$, as well as $I_L$ and $I_H$, in the figure} \xu{add $K_H$ and $I^{NN}_L$})

% \vspace{-5mm}
\noindent{\textbf{Adaptive image resampling operation.}}
Once the interpolation kernels are estimated, they are adaptively applied to the corresponding positions in the LR input image to reconstruct the HR image. 
% Neighboring pixels may be estimated by sampling from the same set of pixels in the LR image space, yet result in different intensity values in the HR image, as each pixel in the HR image space has its own interpolation kernel. 
Nearby pixels in the HR image may be resampled from the same set of pixels in the LR input, yet obtain different intensity values, as each pixel in the HR image space has its own interpolation kernel. 

We first resize the LR input image $I_L$ to the same size of the HR image using the nearest neighbor method, resulting in $I_L^{NN}$. Now $I_L^{NN}$ has the same size as $K_H$ and $I_H$, which is convenient for the implementation of the adaptive resampling (filtering)
operation and further extensions. 
Yet directly applying the interpolation kernels to consecutive elements in $I_L^{NN}$ does not make sense, since neighboring elements in $I_L^{NN}$ include repeated pixels (see Figure~\ref{fig:upscaling}). To apply the estimated kernels to the correct set of pixels within a local region in $I_L$, we need to upscale the interpolation kernels as well, \ie apply them to elements with a certain interval $s$ in $I_L^{NN}$, as shown in Figure~\ref{fig:upscaling} and Equation~\ref{eq:dyn_filt}.
% \hong{mixed usage of dynamic filter and adaptive interpolation kernel, better to use consistent term} \xu{thanks. I have modified the terms to make it consistent.}
%as shown in Equation~\ref{eq:dyn_filt} and Figure~\ref{fig:upscaling}.
%\hong{wrong figure refs. the above explanation can be more concise.}
\begin{equation}
\begin{split}
% 	\hat{I}_H(i,j) = \sum_{{k_1}=0, {k_2}=0}^{f-1, f-1} K_H(&i,j,{k_1}\times f+{k_2}) \times \\
%     I_L^{NN}(&i+s\times({k_1}-f/2), \\
%     &j+s\times({k_2}-f/2)).
	\hat{I}_H(i,j) = &\sum_{{k_1}=0, {k_2}=0}^{f-1, f-1} K_H(i,j,{k_1}\times f+{k_2}) \times \\
    I_L^{NN}&(i+s\times({k_1}-f/2), j+s\times({k_2}-f/2)) .
\end{split}    
\label{eq:dyn_filt}
\end{equation}    
This is similar to the concept of atrous convolution, widely used for semantic segmentation~\cite{Chen-Deeplab,Yu-iclr16}, but is not translation invariant as atrous convolution. The interval $s$ corresponds to the sampling rate parameter in atrous convolution.
Using this scheme, different from traditional interpolation methods, each position in the HR image has a different interpolation kernel which is able to adapt to the appearance and semantics of that position. 
% For a pixel in a smooth region in HR image space, it can be interpolated by simply taking the average of nearby pixels in LR image space. However, for a pixel in a region with strong edge or rich textures, traditional methods, which do not take into account the local context, can not estimate how much a pixel contributes to the reconstruction, making it prone to producing blurry results. 
% TT: left out to gain space:
%We will show how interpolation kernels in different kinds of regions look like in Section~\ref{sec:experiment}.

\subsection{Further Improvements}
\label{sec:refinement}
\noindent{\textbf{Interpolation for larger upscaling factors.}}
Even if the interpolation kernels are estimated with a deep neural network with relatively large receptive fields, the following filtering operation is still a locally linear model.
% \hong{a local linear model$\rightarrow$ locally linear}. 
The elements used for interpolation are limited by the size of the filters. When the upscaling factor gets larger, the correlation between a pixel and its neighbors in the low resolution image becomes smaller. Therefore, the relative performance of interpolation based methods drops as the upscaling factor increases. To reduce this degeneration, we explore two alternatives: i) increasing the size of the interpolation kernels, and ii) doing the upsampling in a progressive way.
\begin{figure}[t]
	\includegraphics[width=0.95\linewidth]{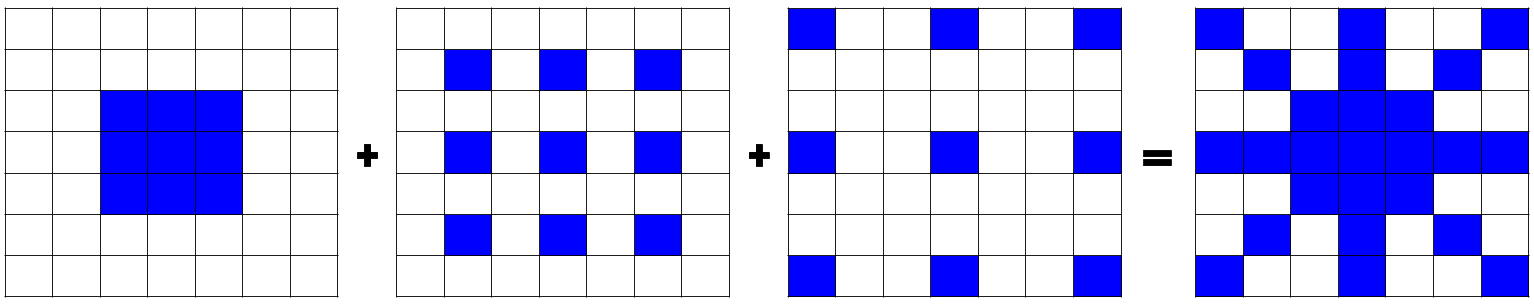}    
    \caption{Demonstration of the atrous spatial pyramid interpolation kernel with filter size 3x3 and interval 1, 2 and 3.}
    \label{fig:asp}
\end{figure}
For the first approach, to sample from a larger neighborhood, a lot more parameters and memory would be required. To alleviate this problem, we borrow the idea of Atrous Spatial Pyramid Pooling (ASPP) from Deeplab-v2 ~\cite{Chen-Deeplab}, which is originally proposed to increase receptive fields for semantic segmentation. 
% Since atrous convolution with different sampling rates can capture different ranges of receptive fields, Chen~\etal proposed to use multiple parallel atrous convolutions with different sampling rates to extract features and fuse them to generate the final result such that the model can capture a large range of receptive fields. 
Similarly, we want the interpolation kernel to cover a larger neighborhood, especially when the upscaling factor is big. 
This can be done by applying the estimated filters to the NN interpolated LR image $I_L^{NN}$ with different intervals $\{s_n\}$, \ie 
\begin{equation}
\begin{split}
%	\hat{I}_H(i,j) = \sum_{{s_n} \in S} \sum_{{k_1}=0, {k_2}=0}^{f-1, f-1} K_H(&i,j,{k_1}\times f+{k_2}) \times \\
%    I_L^{NN}(&i+{s_n}\times({k_1}-f/2), \\
%    &j+{s_n}\times({k_2}-f/2)).
    	\hat{I}_H(i,j) = &\sum_{{s_n} \in S} \sum_{{k_1}=0, {k_2}=0}^{f-1, f-1} K_H(i,j,{k_1}\times f+{k_2}) \times \\
    I_L^{NN}(&i+{s_n}\times({k_1}-f/2), 
    j+{s_n}\times({k_2}-f/2)).
\end{split}    
\label{eq:dyn_filt_aspp}
\end{equation} 
%as is shown in Equation~\ref{eq:dyn_filt_aspp}. 
The sum of the filters over all intervals composes a large interpolation kernel, as is shown in Figure~\ref{fig:asp}. 
The interpolation kernel is sparse but covers a large neighborhood.
% Note that this interpolation kernel is sparse, with many elements equal to zero. 
This way, the range of the local context is enlarged without drastically increasing the number of parameters and memory.

Alternatively, we can also decompose a large upscaling factor into several upsampling operations with smaller upscaling factors. As mentioned in~\cite{Lai-cvpr17}, progressive upsampling makes the super-resolution task easier by dividing it into several sub-problems. Take $4\times$ super-resolution as an example. For progressive upsampling with our model, we first feed a LR image to the model and produce a $2\times$ super-resolved image $\hat{I}_H^{2\times}$. At this stage, it just uses pixels within a local neighborhood to interpolate the pixel in a downsampled version of the high resolution image. At the next stage, we feed $\hat{I}_H^{2\times}$ to another network to estimate another set of interpolation kernels. To avoid drifting away from the content in the original low resolution image, we concatenate the intermediate super-resolution result with the nearest neighbor resized image $I_L^{NN, 2\times}$. The final $4\times$ super-resolved result is obtained by applying the second stage of estimated interpolation kernels to $\hat{I}_H^{2\times}$. \\

\vspace{-2mm}
\noindent{\textbf{Recursive image resampling.}}
Finally, note how the proposed adaptive image resampling approach can be applied several times in a recursive way to further refine the super-resolved result. 
In this case, the initial super-resolved result has already filled in some of the details which were missing in the LR image. 
This intermediate super-resolved result is concatenated with the nearest neighbor resized image and sent to another interpolation kernel estimation module which is used during the recursive process. The estimated interpolation kernels are then applied to the intermediate super-resolved result to refine it further. This process is repeated multiple times with shared parameters for the interpolation kernel estimation modules. This is reminiscent of but different from the recursive layer proposed in~\cite{Kim-cvpr16-drcn} or the recursive block proposed in~\cite{Tai-cvpr2017}. In~\cite{Kim-cvpr16-drcn}, each recursive output estimates a level of residual and the final result is an ensemble of all recursive outputs and the initial bicubic interpolation result. In~\cite{Tai-cvpr2017}, multi-path residual blocks are used to compute a residual and the final result is the sum of the residual and the bicubic interpolation result. 
% In this work, we estimate the interpolation kernels based on the previously super-resolved result and nearest neighbor interpolation result and apply them to the previously result. The whole process is simply repeated for arbitrary times and the network for the recursive interpolation kernel estimation share the same set of weights. 
In our work, the adaptive image resampling is simply repeated multiple times based on the previous super-resolved result. 
% In the recursive process, we share the parameters for the interpolation kernel estimation modules.
%(\hong{not very clear...}\xu{changed. what about current version?})
% TT: rewrote a bit, but still not very clear I'm afraid.
% (\xu{what about now? I move some explanation forward because I think it might be easier to explain together with previous part.})
% The intermediate result is also a reconstruction.

\subsection{Joint Filtering 
%with Adaptive Image Resampling
}
\label{sec:joint}
The proposed adaptive image resampling method is not restricted to the SISR setting. It can be easily extended to joint image filtering tasks such as depth image super-resolution. In this case, the input to the interpolation kernel estimation module includes an additional guidance image $G$ which provides auxiliary information for the filtering process. With help of the guidance image, the model can learn better filters 
%for operations like interpolation 
by considering the content in the guidance image.
Traditional methods proposed in this context, such as joint bilateral upsampling~\cite{Kopf-siggraph07} and guided image filtering~\cite{He-tpami13}, perform spatially-variant filtering operations as well. They compute the output at a pixel as a weighted average of nearby pixels, 
% where the weights are estimated from the guidance image. 
with the weights estimated from the guidance image. 
Instead of hand-designing a function to compute the filter kernel, we use a deep neural network to compute the kernel following a data-driven approach. Deep neural networks are able to model more complex mappings than the hand-designed functions used in~\cite{Kopf-siggraph07,He-tpami13}. 
%\hong{this sentence may be removed.}
Therefore, the proposed model can take full advantage of the guidance image and the filtering input, and integrate them to produce adaptive filters. Different from the SISR setup, here we have $K_H = FCN(I_L, G)$, where $G$ is the guidance image. Then the estimated filters are applied to $I_L^{NN}$ % TT: I_L or I_L^{NN} ? Xu: Changed.
to reconstruct the high resolution image $\hat{I}_H$. This results in sharper edges and less unwanted gradient reversal artefacts.
% (\xu{For the task of depth image super-resolution (maybe describe this in experiment section)})

\section{Experiments}
\label{sec:experiment}
\begin{figure}[t!]
\centering
	\begin{tabular}[t]{c@{\hspace{0.005\linewidth}}c@{\hspace{0.005\linewidth}}c}
		\includegraphics[width=0.32\linewidth]{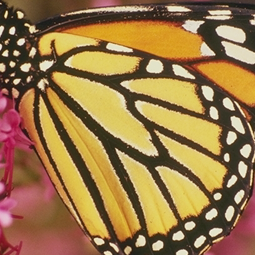}
        &
        \includegraphics[width=0.32\linewidth]{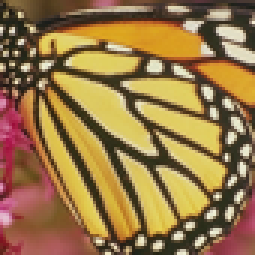}
        &
        \includegraphics[width=0.32\linewidth]{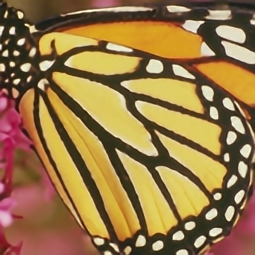}
        \\     
        \footnotesize{(a) HR image}
        &
        \footnotesize{(b) NN resized image}
        &
        \footnotesize{(b) Our result}
        \\          
        \includegraphics[width=0.32\linewidth]{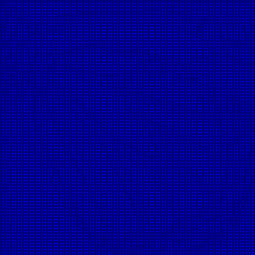}
        &
        \includegraphics[width=0.32\linewidth]{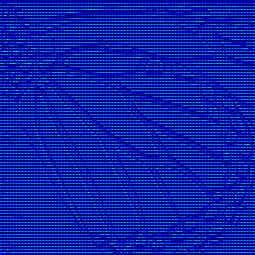}
        &
        \includegraphics[width=0.32\linewidth]{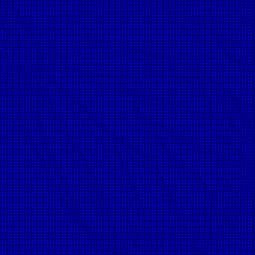}
        \\
        \includegraphics[width=0.32\linewidth]{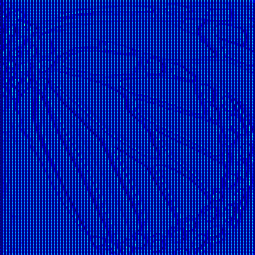}
        &
        \includegraphics[width=0.32\linewidth]{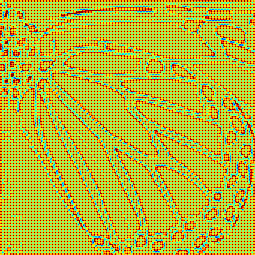}
        &
        \includegraphics[width=0.32\linewidth]{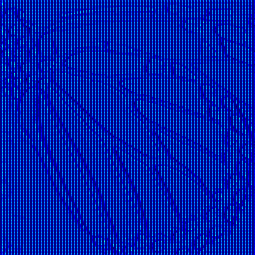}
        \\
        \includegraphics[width=0.32\linewidth]{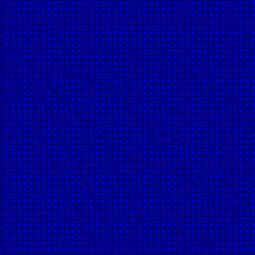}
        &
        \includegraphics[width=0.32\linewidth]{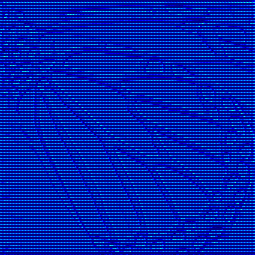}
        &
        \includegraphics[width=0.32\linewidth]{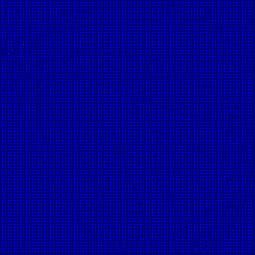}
	\end{tabular}
	\footnotesize{(c) Estimated filters}
    \caption{Visualization of the feature maps corresponding to estimated $5\times 5$ interpolation kernels for $3\times$ super-resolution. Note that we only visualize the inner $3\times 3$ part of the $5\times 5$ kernel since most of the outer part is close to zero. 
%     It is recommended to see (c) in the electronic version. Printed versions may show aliasing artefacts.
	It is recommended to see (c) by zooming in in the electronic version. 
    }
    \label{fig:dyn_filt}
\end{figure}
% \begin{figure}[htp]
% \centering
% 	\begin{tabular}[t]{c@{\hspace{0.005\linewidth}}c@{\hspace{0.005\linewidth}}c}
% 		\includegraphics[width=0.32\linewidth]{./images/filter_visualization/bird_s3_im.png}
%         &
%         \includegraphics[width=0.32\linewidth]{./images/filter_visualization/bird_s3_im_nn.png}
%         &
%         \includegraphics[width=0.32\linewidth]{./images/filter_visualization/bird_s3_out.png}
%         \\     
%         \footnotesize{(a) HR image}
%         &
%         \footnotesize{(b) NN resized image}
%         &
%         \footnotesize{(b) Our result}
% 	\end{tabular}
%     \includegraphics[width=0.95\linewidth]{./images/filter_visualization/bird_s3_filter.png} \\
%     \footnotesize{(c) Filters for each position of the 5x5 filter. \xu{(this is not clear!)}}
%     \caption{Visualization of feature maps correspond to estimated 5x5 interpolation kernels (3x super-resolution). It is recommended to see (c) in the electronic version.}
%     \label{fig:dyn_filt}
% \end{figure}
In this section, we evaluate our models on several widely used single image super-resolution benchmark datasets and visualize the interpolation kernels learned in the proposed model. We also apply it to the joint image filtering task.\\

% dataset and data augmentation
\vspace{-2mm}
\noindent
{\bf Datasets}
We use 291 images as our training data, where 91 images are from Yang~\etal~\cite{Yang-tip10} and 200 images are from the training set of the Berkeley Segmentation Dataset~\cite{Arbelaez-tpami11}. The data is augmented by rotation (90, 180, 270 degrees), scaling (scalefactors of 0.6, 0.7, 0.8, 0.9) and horizontal flipping. Patches of size $96\times 96$, $144\times 144$ and $192\times 192$ are cropped from the augmented data respectively for $2\times$, $3\times$, and $4\times$ super-resolution tasks. 
We downsample the cropped patches to $48\times 48$ using bicubic resizing method. The proposed method is evaluated on four widely used benchmarks: Set5~\cite{Bevilacqua-bmvc12-set5}, Set14~\cite{Zeyde-set14}, BSD100~\cite{Arbelaez-tpami11} and Urban100~\cite{Huang-cvpr15} with SSIM~\cite{Wang-tip04-ssim} and PSNR as measure.\\

\vspace{-2mm}
\noindent
{\bf Implementation details}
As for the FCN used in our interpolation kernel estimation module, we find consecutive standard convolutional layers of $3\times3$ with Relu activation function, but without max-pooling or striding to be effective. Similar to~\cite{Kim-cvpr16-vdsr} the number of filters for all convolutional layers is 64, except for the convolutional layer which produces the adaptive interpolation kernels because that number is dependent on the size of the kernel. 
Unless otherwise mentioned, the interpolation kernel size $f$ is set to 5.
We use subpixel layer~\cite{Shi-cvpr16} as the upsampling layer in this work. 
Simple mean absolute error ($L_1$ loss) is used as our loss function, i.e., $L_1(I_H, \hat{I}_H) = |I_H - \hat{I}_H|$. All models are initialized using the method proposed in~\cite{Glorot-aistats10}
and trained for 200,000 iterations with mini-batches of size 16. Adam optimizer~\cite{Kingma-iclr15-adam} with $\beta_1=0.9$, $\beta_2=0.999$ and $\epsilon=1e-8$ is used to optimize the parameters. The learning rate is initially set to 1e-4 and halved every 50,000 iterations.
% \hong{$\beta$, $\epsilon$?}

\begin{table*}[t!]
\centering
\caption{Ablation study on Set5 and Set14 (PSNR/SSIM).}
% \hong{the caption of table normally appears in the top. }}
\scalebox{0.75}{
	\begin{tabular}{c|c|ccc|c|ccc|cc}
		\toprule
        \multicolumn{2}{c|}{~}
        &
        \multicolumn{3}{c|}{basic model}
        &
        \multicolumn{1}{c|}{baseline}
        &
        \multicolumn{3}{c|}{large upscaling factor}
        &
        \multicolumn{2}{c}{recursive model}
        \\
        \midrule
		Dataset & Scale 
        & DAIR\_5 & DAIR\_10 & DAIR\_20
        & FCN\_20 % Res\_20
        & ASP2\_20 & ASP3\_20 & Prog\_10
        & Recur1\_10 & Recur2\_10
		\\
		\midrule
		~ & $2\times$
		&
		37.42/0.9582 & 37.61/0.9591 & 37.61/0.9592 & 37.43/0.9583 &\textemdash &\textemdash &\textemdash &37.69/0.9594 &37.75/0.9597
        \\
        Set5& $3\times$
		&
		33.47/0.9193 & 33.71/0.9217 & 33.83/0.9228 & 33.57/0.9200 &\textemdash &\textemdash &\textemdash &33.82/0.9230 &33.87/0.9232
        \\
        ~ & $4\times$
		&
		31.19/0.8806 & 31.31/0.8830 & 31.28/0.8828 &31.14/0.8784 &31.27/0.8833 &31.31/0.8844 & 31.47/0.8858 & 31.29/0.8846 &31.43/0.8861
        \\
        \midrule
		~ & $2\times$
		&
		32.92/0.9117 & 33.07/0.9130 & 33.12/0.9131 &32.93/0.9116 &\textemdash &\textemdash &\textemdash &33.16/0.9134 &33.21/0.9139
        \\
        Set14& $3\times$
		&
		29.66/0.8298 & 29.74/0.8317 & 29.78/0.8317 &29.71/0.8302 &\textemdash &\textemdash &\textemdash &29.81/0.8326 &29.85/0.8331
        \\
        ~ & $4\times$
		&
		27.89/0.7651 & 27.97/0.7677 & 27.80/0.7675 &27.83/0.7630 &28.00/0.7673 &27.96/0.7675 &28.04/0.7690 &27.98/0.7681 &28.06/0.7700
        \\
        \bottomrule
	\end{tabular}
}    
\label{tab:ablation_study}
\end{table*}
%%%%%%%%%%%%%%%%%%%%%%%%%%%%%%%%%%%%%%%%%
%\subsection{Model Analysis}
% visualization of learned filters (different positions on one image and different images)
% emphasize the patterns, and difference between smooth regions and regions with textures and boundary
\subsection{Filter visualization}
% (\hong{i comment out and modify some words below, u may check}\xu{I see you comment the inner 3x3. I think it is fine because we mention that in the caption.})
To see how the model exploits the adaptivity
% and spatial awareness 
associated with our image resampling, we take $3\times$ super-resolution as an example and visualize the estimated interpolation kernels -- see Figure~\ref{fig:dyn_filt}.
%Since there is a filter for each position in the HR image space, it is not trivial to visualize the filters for all positions for the whole image. 
Here, instead of directly visualizing the filters at each position, we visualize the feature maps that correspond to the filters. The feature maps have 25 channels where each channel corresponds to one element in a $5\times 5$ filter. 
% Of these, we only show the inner $3\times 3$ block. 
As shown in Figure~\ref{fig:dyn_filt}, the feature map in the middle has higher values than the others. This indicates that the nearest neighbor contributes most to the interpolation, which is consistent with traditional interpolation methods. %like bicubic interpolation and Lanczos resampling. 
% Values on the edges or close to them are lower than the other regions, which implies the reconstruction of edges is more complicated and relies more on information from neighboring points.
The edge regions clearly stand out in those feature maps, indicating that they are treated differently and the interpolation kernels do indeed adapt to the image content.
\begin{figure}[t!]
% \centering
	\begin{tabular}[t]{c@{\hspace{0.005\linewidth}}c@{\hspace{0.005\linewidth}}c}
		\includegraphics[width=0.32\linewidth]{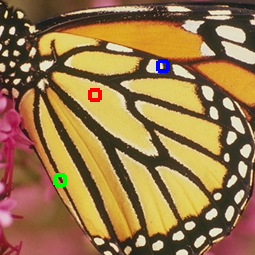}
        &
        \includegraphics[width=0.32\linewidth]{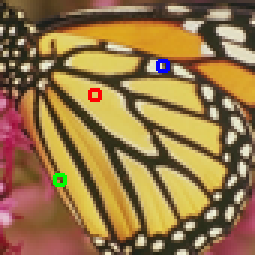}
        &
        \includegraphics[width=0.32\linewidth]{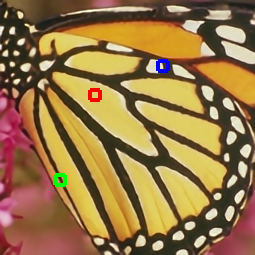}
        \\
        (a) GT
        &
        (b) NN
        &
        (c) Result
    \end{tabular}
    \begin{tabular}[t]{c@{\hspace{0.005\linewidth}}c@{\hspace{0.005\linewidth}}c@{\hspace{0.005\linewidth}}c}
        \includegraphics[width=0.24\linewidth]{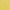}
        &
        \includegraphics[width=0.24\linewidth]{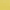}
        &
        \includegraphics[width=0.24\linewidth]{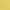}
        &
        \includegraphics[width=0.24\linewidth]{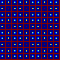}
        \\
        \includegraphics[width=0.24\linewidth]{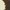}
        &
        \includegraphics[width=0.24\linewidth]{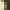}
        &
        \includegraphics[width=0.24\linewidth]{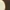}
        &
        \includegraphics[width=0.24\linewidth]{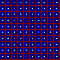}
        \\
        \includegraphics[width=0.24\linewidth]{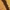}
        &
        \includegraphics[width=0.24\linewidth]{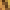}
        &
        \includegraphics[width=0.24\linewidth]{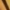}
        &
        \includegraphics[width=0.24\linewidth]{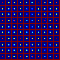}		
        \\
        (d) GT
        &
        (e) NN
        &
        (f) Result
        &
        (g) Filters
	\end{tabular}
    \caption{Visualization of the learned filters for different regions for $3\times$ super-resolution. The rows from top to bottom correspond to the red, blue and green regions respectively. Each cell in the grids in (g) corresponds to a $5\times 5$ filter. It is recommended to zoom in for details.}
    \label{fig:dyn_filt_part}
\end{figure}
Further, we can see that
% the feature maps next to the middle one
the middle feature map and the ones next to it
show certain patterns, that is, vertical stripes for its left and right neighbors and horizontal stripes for its top and bottom neighbors. 
This reflects the variation due to the relative location of the HR pixel to its nearest region in the LR image.
The elements next to the nearest neighbor complement with each other to obtain the best combination, especially on the edges. This increases the contrast between two sides of an edge such that the edge looks sharper.

% \xu{we are not sure that is checkerboard artefact, so commented}
% We also observe that the feature maps show some checkerboard artefacts, which is mainly caused by the upsampling layer and is also reported in~\cite{Odena-distill16}. However, we do not observe any remaining checkerboard artefacts in the super-resolved result. Our explanation is that though the interpolation kernels have checkerboard artefacts, they are applied to different elements in the low resolution image. Therefore, the checkerboard artefacts are alleviated by the adaptive filtering operation.

To see how the filters adapt to different regions, we show some example filters that correspond to a smooth region
 , a textured region 
%  (\hong{why comment out?} \xu{now we do not have a textured region to show because the region is too small to cover many textures})
and a region with strong edge in Figure~\ref{fig:dyn_filt_part}. Each cell in the grid in column (g) corresponds to a $5\times 5$ filter. We find that for a smooth region the distribution of filters follows a regular pattern, that is, there are several patterns at different positions and one pattern repeatedly appears with an interval of the upscaling factor 3 (note that the values of the filters are not exactly the same but very close) -- for example, the grid with dark red dot in the center and its 8 neighbors. It is easy to understand this because pixels at different positions in a smooth region of a low resolution image look similar and they can be handled in a similar way for high resolution reconstruction.
However, this is not the case for regions with rich textures and strong edges. Filters in such regions do not follow a regular pattern but become more complicated. Spatially invariant interpolation kernels do not work well. The kernels need to adapt to the variance within a region such as the edge and texture.
\subsection{Ablation Study}
In this section, we study the effects of different components of our method described in Section~\ref{sec:method}. \\

% ablation study
\vspace{-2mm}
\noindent{\textbf{Number of layers.}}
% different number of layers in filter generation network;
First, we study the effect of the number of layers used to estimate the adaptive interpolation kernels. We experiment with 5, 10 and 20 layers for the adaptive interpolation kernel module, referred to as DAIR$\_5$, DAIR$\_10$ and DAIR$\_20$. Table~\ref{tab:ablation_study} shows that the deeper the network for that module, the better the super-resolved result it obtains. However, the difference between models with 10 and 20 convolutional layers is small. To  have a compact and effective model, we use 10 convolutional layers for each stage for our progressive and recursive models (see below). \\

% w/ and w/o dfn;
\vspace{-2mm}
\noindent{\textbf{Adaptive image resampling.}}
We compare the proposed model and a model without adaptive image resampling operation, FCN\_20. The architecture is similar to the network architecture in SRCNN~\cite{Dong-tpami16} in that it is simply a fully convolutional nework, but different from it in two points: it starts from a nearest neighbor resized image instead of a bicubic interpolation resized one; and it has the same number of convolutional layers as DAIR\_20 which is more than that in SRCNN. 
\begin{table*}[tp]
\centering
\caption{Quantitative results on benchmark datasets (PSNR/SSIM). The best is marked with \textcolor{red}{red} and the second best is marked with \textcolor{blue}{blue}.}
% \hong{also indicate the first and second best values?}}
\scalebox{0.75}{
	\begin{tabular}{ccccccccccc}
		\toprule
		Dataset & Scale & Bicubic &Lanczos3 
        & SRCNN~\cite{Dong-tpami16} & FSRCNN~\cite{Dong-eccv16} & VDSR~\cite{Kim-cvpr16-vdsr} & DRCN~\cite{Kim-cvpr16-drcn} & LapSRN~\cite{Lai-cvpr17} &DRRN~\cite{Tai-cvpr2017}
        & Ours
		\\
		\midrule
		~ & $2\times$
		&33.66/0.9299 &34.32/0.9365
        &36.66/0.9542 &37.05/0.956 &37.53/0.9587 &37.63/0.9588 &37.52/0.959 &\textcolor{blue}{37.74}/\textcolor{blue}{0.9591}
        &\textcolor{red}{37.75}/\textcolor{red}{0.9597}
        \\
        Set5& $3\times$
		&30.89/00.8682 &30.82/0.8754
        &32.75/0.9090 &33.18/0.914 &33.66/0.9213 &33.82/0.9226 &\textemdash &\textcolor{red}{34.03}/\textcolor{red}{0.9244}
        &\textcolor{blue}{33.87}/\textcolor{blue}{0.9232}
        \\
        ~ & $4\times$
		&28.42/0.8104 &28.80/0.8178
        &30.48/0.8628 &30.72/0.866 &31.35/0.8838 &31.53/0.8854 &\textcolor{blue}{31.54}/0.885 &\textcolor{red}{31.68}/\textcolor{red}{0.8888}
        &31.47/\textcolor{blue}{0.8865}
        \\
        \midrule
		~ & $2\times$
		&30.24/0.8688 &30.69/0.8791
        &32.45/0.9067 &32.66/0.909 &33.03/0.9124 &33.04/0.9118 &33.08/0.913 &\textcolor{red}{33.23}/\textcolor{blue}{0.9136}
        &\textcolor{blue}{33.21}/\textcolor{red}{0.9139}
        \\
        Set14& $3\times$
		&27.55/0.7742 &27.83/0.7830
        &29.30/0.8215 &29.37/0.824 &29.77/0.8314 &29.76/0.8311 &\textemdash &\textcolor{red}{29.96}/\textcolor{red}{0.8349}
        &\textcolor{blue}{29.85}/\textcolor{blue}{0.8331}
        \\
        ~ & $4\times$
		&26.00/0.7027 &26.23/0.7098
        &27.50/0.7513 &27.61/0.755 &28.01/0.7674 &28.02/0.7670 &\textcolor{blue}{28.19}/\textcolor{red}{0.772} &\textcolor{red}{28.21}/\textcolor{red}{0.7720}
        &28.07/\textcolor{blue}{0.7701}
        \\
        \midrule
		~ & $2\times$
		&29.56/0.8431 &29.92/0.8551
        &31.36/0.8879 &31.53/0.892 &31.90/0.8960 &31.85/0.8942 &31.80/0.895 &\textcolor{red}{32.05}/\textcolor{blue}{0.8973}
        &\textcolor{blue}{32.00}/\textcolor{red}{0.8974}
        \\
        BSD100& $3\times$
		&27.21/0.7385 &27.41/0.7481
        &28.41/0.7863 &28.53/0.791 &28.82/0.7976 &28.80/0.7963 &\textemdash &\textcolor{red}{28.95}/\textcolor{red}{0.8004}
        &\textcolor{blue}{28.87}/\textcolor{blue}{0.7991}
        \\
        ~ & $4\times$
		&25.96/0.6675 &26.13/0.6754
        &26.90/0.7101 &26.98/0.715 &27.29/0.7251 &27.23/0.7233 &\textcolor{blue}{27.32}/\textcolor{blue}{0.728} &\textcolor{red}{27.38}/\textcolor{red}{0.7284}
        &27.25/0.7263
        \\
        \midrule
		~ & $2\times$
		&26.88/0.8403 &27.25/0.8503
        &29.50/0.8946 &29.88/0.902 &30.76/0.9140 &30.75/0.9133 &30.41/0.910 &\textcolor{red}{31.23}/\textcolor{red}{0.9188}
        &\textcolor{blue}{31.08}/\textcolor{blue}{0.9176}
        \\
        Urban100& $3\times$
		&24.46/0.7349 &24.68/0.7430
        &26.24/0.7989 &26.43/0.808 &27.14/0.8279 &27.15/0.8276 &\textemdash &\textcolor{red}{27.53}/\textcolor{red}{0.8378}
        &\textcolor{blue}{27.24}/\textcolor{blue}{0.8317}
        \\
        ~ & $4\times$
		&23.14/0.6577 &23.32/0.6641
        &24.52/0.7221 &24.62/0.728 &25.18/0.7524 &25.14/0.7510 &\textcolor{blue}{25.21}/\textcolor{blue}{0.756} &\textcolor{red}{25.44}/\textcolor{red}{0.7638}
        &25.13/0.7549
        \\
        \bottomrule
	\end{tabular}
}    
\label{tab:comparison_sota}
\end{table*}
\begin{figure*}	 
    \begin{tabular}[t]{c@{\hspace{0.005\linewidth}}c@{\hspace{0.005\linewidth}}c@{\hspace{0.005\linewidth}}c@{\hspace{0.005\linewidth}}c@{\hspace{0.005\linewidth}}c@{\hspace{0.005\linewidth}}c}
    Ground Truth
    &
    Bicubic
    &
    SRCNN
    &
    VDSR
    &
    DRCN
    &
    LapSRN
    &
    Ours
    \\
    \includegraphics[width=0.14\linewidth]{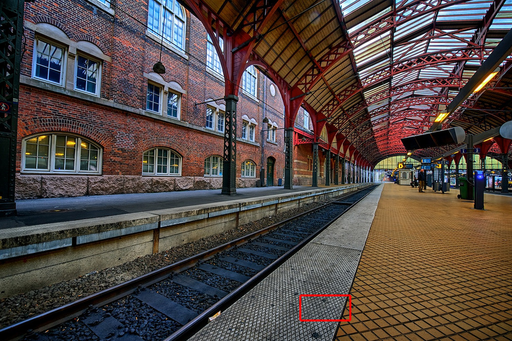}
    &
    \includegraphics[width=0.14\linewidth]{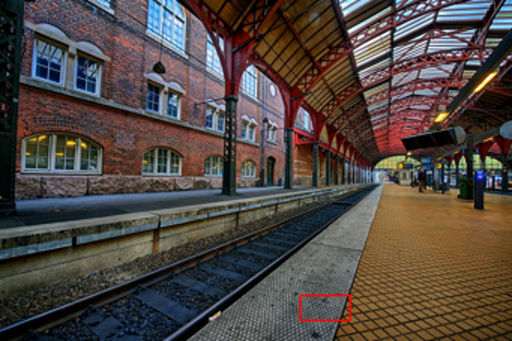}
    &
    \includegraphics[width=0.14\linewidth]{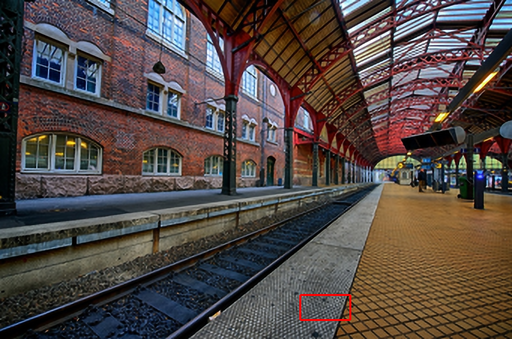}
    &
    \includegraphics[width=0.14\linewidth]{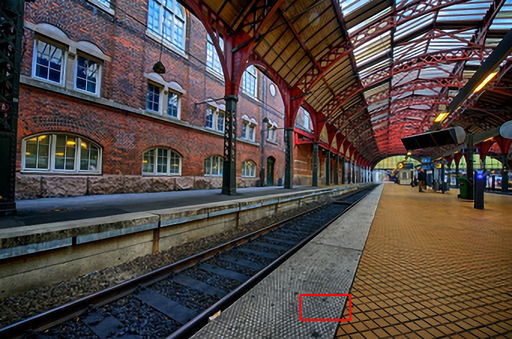}
    &
    \includegraphics[width=0.14\linewidth]{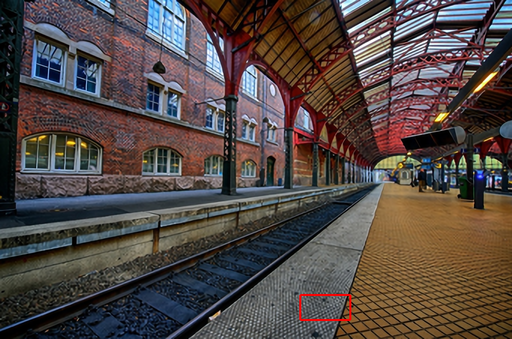}
    &
    \includegraphics[width=0.14\linewidth]{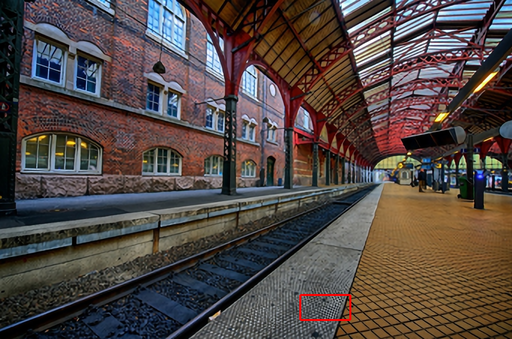}
    &
    \includegraphics[width=0.14\linewidth]{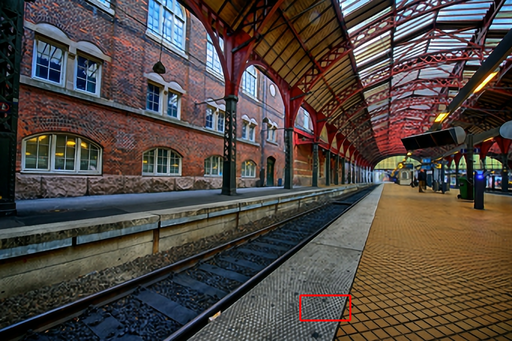}
    \\
    \includegraphics[width=0.14\linewidth]{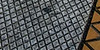}
    &
    \includegraphics[width=0.14\linewidth]{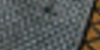}
    &
    \includegraphics[width=0.14\linewidth]{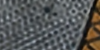}
    &
    \includegraphics[width=0.14\linewidth]{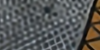}
    &
    \includegraphics[width=0.14\linewidth]{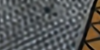}
    &
    \includegraphics[width=0.14\linewidth]{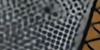}
    &
    \includegraphics[width=0.14\linewidth]{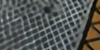} 
    \\
    (PSNR/SSIM)
    &
    (20.43/0.4914)    
    &
    (21.06/0.5735)
    &
    (21.34/0.6034)
    &
    (21.36/0.6025)
    &
    (21.28/0.6018)
    &
	(21.30/0.6046)
    \\
    \includegraphics[width=0.14\linewidth]{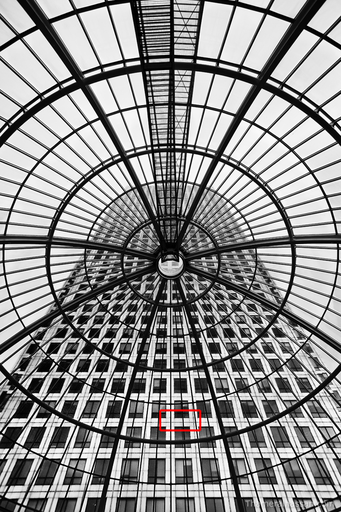}
    &
    \includegraphics[width=0.14\linewidth]{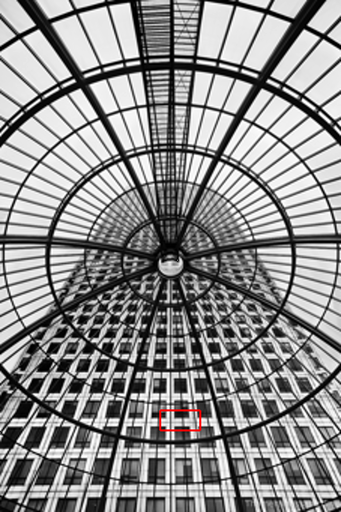}
    &
    \includegraphics[width=0.14\linewidth]{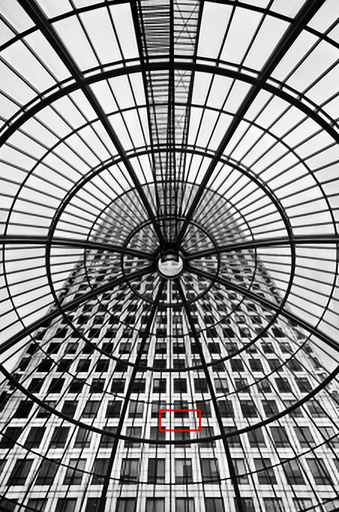}
    &
    \includegraphics[width=0.14\linewidth]{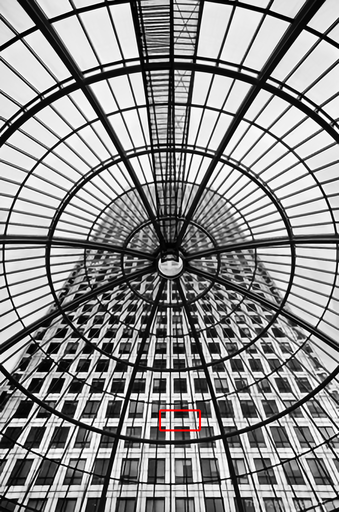}
    &
    \includegraphics[width=0.14\linewidth]{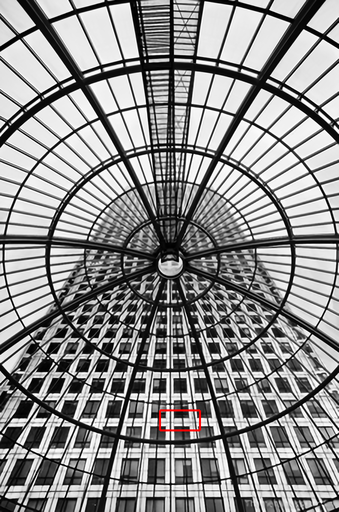}
    &
    \includegraphics[width=0.14\linewidth]{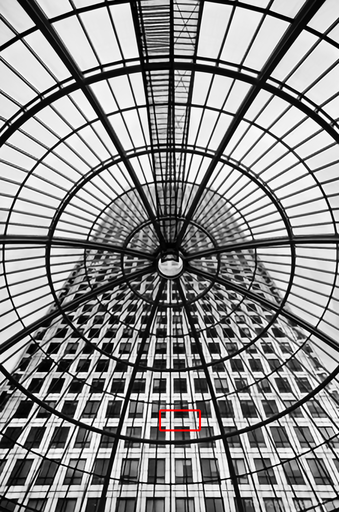}
    &
    \includegraphics[width=0.14\linewidth]{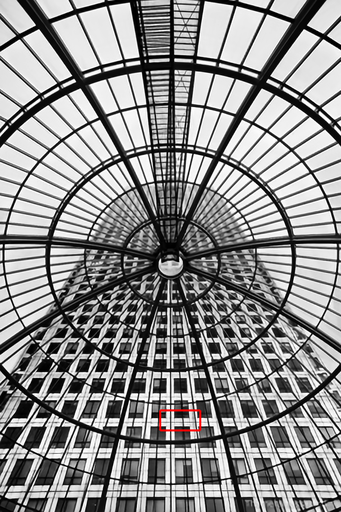}
    \\
    \includegraphics[width=0.14\linewidth]{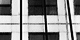}
    &
    \includegraphics[width=0.14\linewidth]{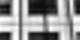}
    &
    \includegraphics[width=0.14\linewidth]{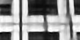}
    &
    \includegraphics[width=0.14\linewidth]{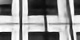}
    &
    \includegraphics[width=0.14\linewidth]{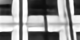}
    &
    \includegraphics[width=0.14\linewidth]{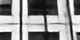}
    &
    \includegraphics[width=0.14\linewidth]{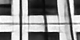} 
    \\
    (PSNR/SSIM)
    &
    (18.07/0.6839)
    &
    (21.27/0.8130)
    &
    (21.22/0.8629)
    &
    (21.25/0.8621)
    &
    (21.18/0.8624)
    &
	(21.31/0.8649)
    \\
    \end{tabular}
    \caption{\label{fig:qualitative_sota}
    Qualitative comparison on 3x single image super-resolution.}
    \vspace{-3mm}
\end{figure*}
The numbers in Table~\ref{tab:ablation_study} show that the method with adaptive resampling module is much better than the one without. Instead of directly generating a new value as a pixel in HR image space, we propose a way to interpolate nearby pixels in LR image space. This shares a similar philosophy as the global residual connection in the last layer of~\cite{Kim-cvpr16-vdsr,Kim-cvpr16-drcn}, which makes the training easier and has been proven to improve the performance. \\

\vspace{-2mm}
\noindent{\textbf{Large upscaling factor.}}
% atrous spatial pyramid; progressive
As is mentioned in Section~\ref{sec:method}, we explored two ways to address the weakness of interpolation-based methods in case of large upscaling factors. One way is to use the sum of an atrous spatial pyramid to approximate a large interpolation kernel such that it can combine information from a large neighborhood. We experimented with 2-level and 3-level atrous spatial pyramids, denoted as ASP$_2$ and ASP$_3$ in Table~\ref{tab:ablation_study}. 
% We find that the method with a 3-level atrous spatial pyramid obtains better performance because its approximated kernels can cover larger context. 
We find that the use of atrous spatial pyramid somewhat improves the performance because its approximated kernels can cover larger context. 
The alternative way %approach 
we proposed is progressive upsampling, i.e. to progressively upsample a low-resolution image to an intermediate resolution image and further upsample it to a high-resolution one. The result shows progressive upsampling significantly improves the result in case of $4\times$ super-resolution. We conclude that the progressive upsampling for large upscaling factor is more effective and this will be used in our final model for $4\times$ super-resolution. 

\noindent{\textbf{Recursive refinement.}}
We first apply a basic deep adaptive image resampling module with 10 convolutional layers to the low resolution image and obtain an initial result. Then another adaptive image resampling module with 10 convolutional layers is recursively applied to the previous result. We experiment with one time (Recur1\_10) and two times (Recur2\_10) recursion. The number of parameters for Recur1\_10 and Recur2\_10 are the same and the difference is the times the adaptive image resampling is applied. Table~\ref{tab:ablation_study} shows that more iterations further refines the super-resolution result. We will use Recur2\_10 in our final model.
% for $2\times$ and $3\times$ super-resolution.

\begin{figure*}[t]
% \centering
	\begin{tabular}[t]{c@{\hspace{0.005\linewidth}}c@{\hspace{0.005\linewidth}}c@{\hspace{0.005\linewidth}}c@{\hspace{0.005\linewidth}}c}
		\includegraphics[width=0.19\linewidth]{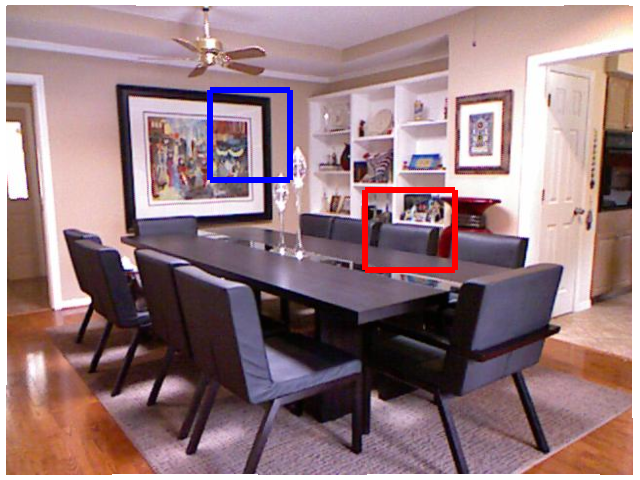}
        &
        \includegraphics[width=0.19\linewidth]{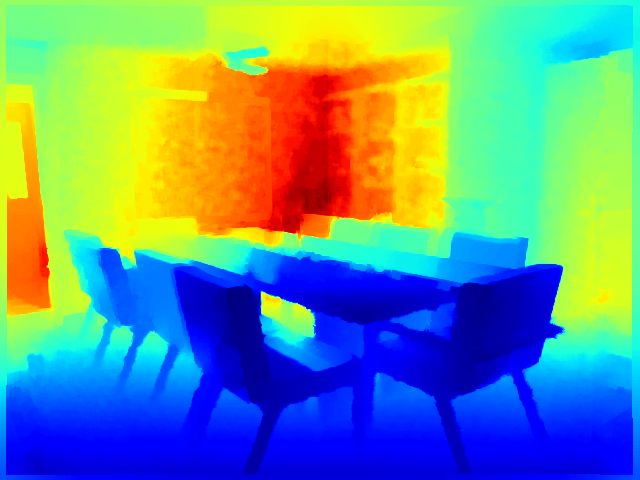}
        &
        \includegraphics[width=0.19\linewidth]{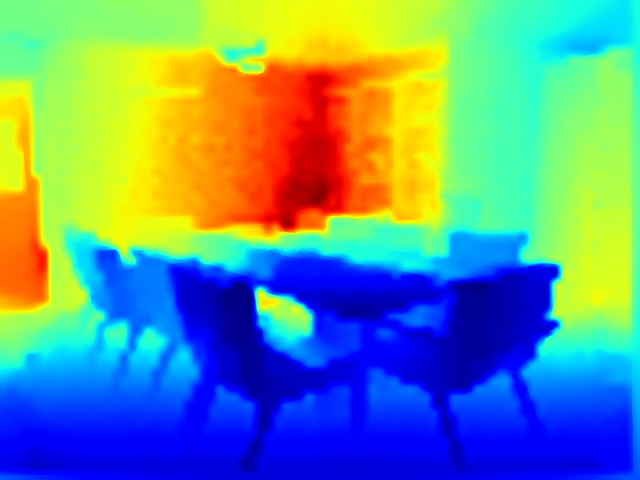}
        &
        \includegraphics[width=0.19\linewidth]{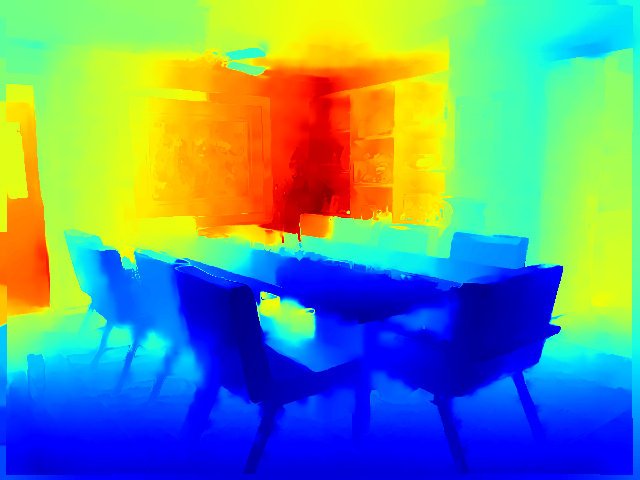}
        &
        \includegraphics[width=0.19\linewidth]{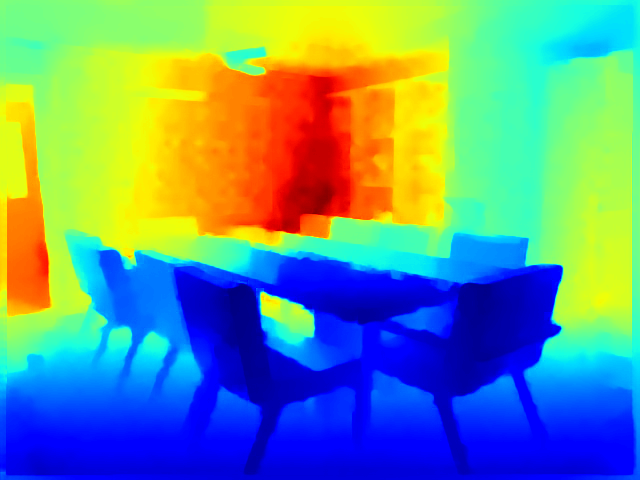}
        \\
        \begin{tabular}[t]{c@{\hspace{0.005\linewidth}}c}
          \includegraphics[width=0.08\linewidth, height=0.08\linewidth]{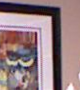}
          &
          \includegraphics[width=0.08\linewidth, height=0.08\linewidth]{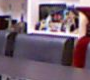}
        \end{tabular}
        &
        \begin{tabular}[t]{c@{\hspace{0.005\linewidth}}c}
          \includegraphics[width=0.08\linewidth, height=0.08\linewidth]{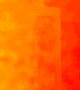}
          &
          \includegraphics[width=0.08\linewidth, height=0.08\linewidth]{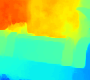}
        \end{tabular}
        &
        \begin{tabular}[t]{c@{\hspace{0.005\linewidth}}c}
          \includegraphics[width=0.08\linewidth, height=0.08\linewidth]{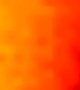}
          &
          \includegraphics[width=0.08\linewidth, height=0.08\linewidth]{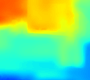}
        \end{tabular}
        &
        \begin{tabular}[t]{c@{\hspace{0.005\linewidth}}c}
          \includegraphics[width=0.08\linewidth, height=0.08\linewidth]{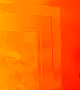}
          &
          \includegraphics[width=0.08\linewidth, height=0.08\linewidth]{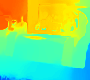}
        \end{tabular}
        &
        \begin{tabular}[t]{c@{\hspace{0.005\linewidth}}c}
          \includegraphics[width=0.08\linewidth, height=0.08\linewidth]{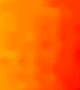}
          &
          \includegraphics[width=0.08\linewidth, height=0.08\linewidth]{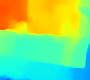}
        \end{tabular}        
        \\
        \small
        \small{(a) Guidance}
        &
        \small{(b) GT}
        &
        \small{(c) GF}
        &
        \small{(d) JBU}
        &
        \small{(e) Ours}
    \end{tabular}
    \caption{\label{fig:joint_filter} Qualitative comparison with JBU and GF ($8\times$ upsampling) on depth map super-resolution.}
\end{figure*}
\subsection{Comparison with state-of-the-art methods}
% number of parameters; maybe also time on titan-x pascal
% analyze why it performs better on 2x and 3x but a little worse on 4x;
% and also mention that it is complementary with other state-of-the-art methods, which can further refined by our method.
% (\xu{todo: final model, esp. for $4\times$})
We use Recur2\_10 as our final model for $2\times$ and $3\times$ super-resolution and combine Prog\_10 and Recur2\_10 as the final model for $4\times$ super-resolution.
We compare the proposed deep adaptive image resampling method with several state-of-the-art methods: SRCNN~\cite{Dong-tpami16}, FSRCNN~\cite{Dong-eccv16}, VDSR~\cite{Kim-cvpr16-vdsr}, DRCN~\cite{Kim-cvpr16-drcn}, LapSRN~\cite{Lai-cvpr17} and DRRN~\cite{Tai-cvpr2017}. As shown in Table~\ref{tab:comparison_sota},
% \hong{wrong table ref}
our method achieves competitive performance to state-of-the-art methods on four benchmarks, especially for small upscaling factors. 
% Since PSNR and SSIM are not completely correlated with human perception of image super-resolution, we also show some visual comparison. 
We also show some visual comparison results.
In Figure~\ref{fig:qualitative_sota}, we can find that the proposed method performs well on textured regions and regions with strong edges.

\subsection{Joint Image Filtering}
To demonstrate the effectiveness of the proposed model in joint image filtering, we carry out an experiment on the task of depth map super-resolution. The basic model with 10 convolutional layers, i.e. DAIR\_10, is used here to validate the effectiveness of the deep adaptive image resampling on joint image filtering. 
% The input to the filter generation network is the combination of a downsampled depth map and a guidance RGB image, and the target input is the downsampled depth map. 
A downsampled depth map is first resized using nearest neighbor interpolation and then concatenated with a guidance RGB image as the input to the interpolation kernel generation module.
The high resolution reconstruction is computed by applying the estimated adaptive interpolation kernels to the nearest neighbor interpolation of the downsampled depth map.

Similar to~\cite{Li-eccv16-DJF}, we also collect the training data by cropping patch pairs from 1,000 RGB images and depth maps in the NYU-v2 dataset~\cite{Silberman-eccv12-nyuv2}. The sizes of cropped patches are $64 \times 64$ for $4\times$ and $8\times$ and $128 \times 128$ for $16\times$ upsampling. The depth map patches are downsampled using the nearest neighbor method as low-resolution input and the RGB patches are used as guidance. Once the model is trained, it is evaluated on two datasets, the rest of 449 images in NYU-v2 dataset and Middlebury dataset~\cite{middleburry-cvpr07a,middleburry-cvpr07b} with missing values filled in by Lu~\etal~\cite{middleburry-cvpr14}. 

% \xu{visualize which kind of filters are learned; or put in the supplementary material}
% comparison with sota methods (qualitative and quantitative)
We compare the proposed method with several joint image filtering methods, which include JBU~\cite{Kopf-siggraph07}, GF~\cite{He-tpami13}, TGV~\cite{Ferstl-iccv13}, MSG-Net~\cite{Hui-eccv16}, FBS~\cite{Barron-eccv16} and DJF~\cite{Li-eccv16-DJF}. Quantitative results\footnote{The results of other methods are obtained from~\cite{Li-eccv16-DJF}.} are shown in Table~\ref{tab:depth_superres} 
% \hong{wrong table ref} 
with root mean squared errors (RMSE) as evaluation metric.
\begin{table}[htbp]
\centering
\caption{\label{tab:depth_superres} Quantitative comparison on depth map super-resolution. 
The best is marked with \textcolor{red}{red} and the second best with \textcolor{blue}{blue}.}
% Numbers in \textcolor{red}{red} indicate the best performance and the ones in \textcolor{blue}{blue} indicate the second best.
\vspace{2mm}
\scalebox{0.9}
  {
  \begin{tabular}{l|ccc|ccc}
  \toprule
  % ~ & \textbf{Middlebury \cite{middleburry-cvpr07a,middleburry-cvpr07b}} & \textbf{NYU-v2 \cite{Silberman-eccv12-nyuv2}} \\
  % \multicolumn{1}{c}{} & {4$\times$~~ ~8$\times$~~ 16$\times$} & {4$\times$~~ ~8$\times$~~ 16$\times$}\\
  ~ & \multicolumn{3}{c|}{\textbf{Middlebury~\cite{middleburry-cvpr07a,middleburry-cvpr07b}}}
  &
  \multicolumn{3}{c}{\textbf{NYU-v2 \cite{Silberman-eccv12-nyuv2}}}
  \\
  ~ & 4$\times$ & 8$\times$ & 16$\times$ & 4$\times$ & 8$\times$ & 16$\times$
  \\
  \midrule
  Bicubic &4.44 &7.58 &11.87 &8.16 &14.22 &22.32 \\
  GF \cite{He-tpami13} &4.01 &7.22 &11.70 &7.32 &13.62 &22.03 \\
  JBU \cite{Kopf-siggraph07} &2.44 &3.81 &6.13 &4.07 &8.29 &13.35 \\
  TGV \cite{Ferstl-iccv13} &3.39 &5.41 &12.03 &6.98 &11.23 &28.13 \\

  MSG-Net \cite{Hui-eccv16} &\textcolor{red}{1.79} &\textcolor{blue}{3.39} &\textcolor{red}{5.87} &3.78 &6.37 &11.16 \\
  FBS \cite{Barron-eccv16} &2.58 &4.19 &7.30 &4.29 &8.94 &14.59 \\
  DJF~\cite{Li-eccv16-DJF} &2.14 &3.77 &6.12 &\textcolor{blue}{3.54} &\textcolor{blue}{6.20} &\textcolor{blue}{10.21} \\
  Ours &\textcolor{red}{1.79} &\textcolor{red}{3.27} &\textcolor{blue}{6.08} &\textcolor{red}{2.67} &\textcolor{red}{5.86} &\textcolor{red}{10.03} \\
  \bottomrule
  \end{tabular}
  }
\end{table}
We can see that the proposed method performs favorably against state-of-the-art methods. Compared with~\cite{Hui-eccv16}, which uses a multi-scale guidance, our method is simpler and more effective. Besides, our method which is trained on NYU-v2 dataset also shows good generalization ability when evaluated on Middleburry dataset. We attribute this to the adaptive nature of the proposed method. Similar to the experiment on SISR, we observe that the performance somewhat decreases in case of large upscaling factor, especially for $16\times$ upsampling. 
% That is because our method can only interpolate nearby pixels. When nearby pixels have little correlation, our method will fail to reconstruct the high resolution pixel. 
%While for other methods they are equipped with the ability to directly predict the depth map from the guidance image.
For large upscaling factors, other deep learning based methods may learn to directly predict the depth map from the guidance image, which is impossible with our approach.
% \hong{still not clear why your method performs well at large upscaling factor}\xu{the reason why our method works is still because of the adaptive interpolation}
% TT: Reformulated the above. Yet it sounds quite defensive, given that we're still best on NYU and second best on Middlebury. Xu: Agree.
We also show a qualitative comparison with filter-based methods in Figure~\ref{fig:joint_filter}.
As shown, JBU and GF, both based on hand-designed spatially variant filters, can adapt to the content of the image to some extent. However, they tend to pay much attention to the strong edges even if the two sides of the edge have the same depth. They are also prone to produce artefacts on the boundary. 
%Our method estimates the interpolation kernels with a deep neural network in a data-driven fashion. 
The deep neural network used in our model is able to capture more complicated relations among LR input, guidance and HR output. 
% Therefore, 
Hence the proposed method is more powerful than the methods based on hand-designed spatially variant filters.  
% \xu{mention the speed (CPU and on Titan x-pascal)}
% our method is 4.19s on Intel i7 3.5GHz CPU, and 0.08s on titan X pascal GPU on images of size 640x480, the same for 8x

\section{Conclusion}
In this paper we propose a Deep Adaptive Image Resampling method to address the image super-resolution task. Spatially variant interpolation kernels are estimated with a convolutional neural network and then applied to a low resolution image to reconstruct the high resolution image. We demonstrate the effectiveness of the proposed method by evaluating it on both single image super-resolution and joint image filtering tasks. Visualization of the estimated inteporlation kernels gives more insight on the effectiveness of the proposed method.

{\small
\bibliographystyle{ieee}
\bibliography{main}
}

\end{document}

%%%%%%%%% TITLE
\title{Super-Resolution with Deep Adaptive Image Resampling}
% Single Image Super-Resolution with Deep Adaptive Image Resampling
% Revisiting resampling method for single image super-resolution with deep learning

\author{First Author\\
Institution1\\
Institution1 address\\
{\tt\small firstauthor@i1.org}
\and
Second Author\\
Institution2\\
First line of institution2 address\\
{\tt\small secondauthor@i2.org}
}

\maketitle
%\thispagestyle{empty}

%%%%%%%%% BODY TEXT
\section{Network architecture}
In this section, we describe the network architecture of the proposed models in Table~\ref{tab:network_dair_10} and Table~\ref{tab:network_recur2_10}. The subpixel layer we used is the one proposed in~\cite{Shi-cvpr16}.
\begin{table}[hp]
\caption{Network architecture of our basic model DAIR\_10.}
\begin{center}
\scalebox{0.6}{
\begin{tabular}{|c|c|c|c|}
\hline
 input & layer & param & output \\
\hline 
 $w\times h\times 1$ & conv & $1\times3\times3\times 64$ & $w\times h\times 64$ \\
 $w\times h\times 64$ & conv & $64\times3\times3\times 64$ & $w\times h\times 64$ \\
 $w\times h\times 64$ & conv & $64\times3\times3\times 64$ & $w\times h\times 64$ \\
 $w\times h\times 64$ & conv & $64\times3\times3\times 64$ & $w\times h\times 64$ \\
 $w\times h\times 64$ & conv & $64\times3\times3\times 64$ & $w\times h\times 64$ \\
 $w\times h\times 64$ & conv & $64\times3\times3\times 64$ & $w\times h\times 64$ \\
 $w\times h\times 64$ & conv & $64\times3\times3\times 64$ & $w\times h\times 64$ \\
 $w\times h\times 64$ & conv & $64\times3\times3\times 64$ & $w\times h\times 64$ \\
 $w\times h\times 64$ & conv & $64\times3\times3\times 64$ & $w\times h\times 64$ \\
 $w\times h\times 64$ & conv & $64\times3\times3\times 64$ & $w\times h\times 64$ \\
 $w\times h\times 64$ & conv & $64\times3\times3\times (64\times s^2)$ & $w\times h\times (64\times s^2)$ \\
 $w\times h\times (64\times s^2)$ & subpixel & ~ & $(w\times s)\times (h\times s) \times 64$ \\
 $(w\times s)\times (h\times s) \times 64$ & conv & $64\times3\times3\times 25$ & $(w\times s)\times (h\times s)\times 25$ \\
 $w\times h\times 1$ &NN interp &~ &$(w\times s) \times (h\times s) \times 1$ \\
\hline 
 $(w\times s) \times (h\times s) \times 1 $ & adaptive &~ &~ \\
 $\&$ & image &~ &$(w\times s) \times (h\times s) \times 1$ \\
$(w\times s)\times (h\times s) \times 25$ & resampling &~ &~ \\
\hline
\end{tabular}
}
\end{center}
\label{tab:network_dair_10}
\end{table}
% Recursive 2 times
\begin{table}[tp]
\caption{Network architecture of our basic model Recur2\_10.}
\begin{center}
\scalebox{0.6}{
\begin{tabular}{|c|c|c|c|}
\hline
 input & layer & param & output \\
\hline 
 $w\times h\times 1$ & conv & $1\times3\times3\times 64$ & $w\times h\times 64$ \\
 $w\times h\times 64$ & conv & $64\times3\times3\times 64$ & $w\times h\times 64$ \\
 $w\times h\times 64$ & conv & $64\times3\times3\times 64$ & $w\times h\times 64$ \\
 $w\times h\times 64$ & conv & $64\times3\times3\times 64$ & $w\times h\times 64$ \\
 $w\times h\times 64$ & conv & $64\times3\times3\times 64$ & $w\times h\times 64$ \\
 $w\times h\times 64$ & conv & $64\times3\times3\times 64$ & $w\times h\times 64$ \\
 $w\times h\times 64$ & conv & $64\times3\times3\times 64$ & $w\times h\times 64$ \\
 $w\times h\times 64$ & conv & $64\times3\times3\times 64$ & $w\times h\times 64$ \\
 $w\times h\times 64$ & conv & $64\times3\times3\times 64$ & $w\times h\times 64$ \\
 $w\times h\times 64$ & conv & $64\times3\times3\times 64$ & $w\times h\times 64$ \\
 $w\times h\times 64$ & conv & $64\times3\times3\times (64\times s^2)$ & $w\times h\times (64\times s^2)$ \\
 $w\times h\times (64\times s^2)$ & subpixel & ~ & $(w\times s)\times (h\times s) \times 64$ \\
 $(w\times s)\times (h\times s) \times 64$ & conv & $64\times3\times3\times 25$ & $(w\times s)\times (h\times s)\times 25$ \\
 $w\times h\times 1$ &NN interp &~ &$(w\times s) \times (h\times s) \times 1$ \\
\hline 
 $(w\times s) \times (h\times s) \times 1$ & adaptive &~ &~ \\
 $\&$ & image &~ &$(w\times s) \times (h\times s) \times 1$ \\
 $(w\times s)\times (h\times s) \times 25$ & resampling &~ &~ \\
\hline
 $(w\times s) \times (h\times s) \times 1$ & concat &~ & $(w\times s) \times (h\times s) \times 2$ \\
 $(w\times s) \times (h\times s) \times 1$ &~ &~ &~ \\
\hline 
 $(w\times s) \times (h\times s) \times 2$ & conv & $2\times3\times3\times 64$ & $(w\times s) \times (h\times s) \times 64$ \\
 $(w\times s) \times (h\times s) \times 64$ & conv & $64\times3\times3\times 64$ & $(w\times s) \times (h\times s) \times 64$ \\
 $(w\times s) \times (h\times s) \times 64$ & conv & $64\times3\times3\times 64$ & $(w\times s) \times (h\times s) \times 64$ \\
 $(w\times s) \times (h\times s) \times 64$ & conv & $64\times3\times3\times 64$ & $(w\times s) \times (h\times s) \times 64$ \\
 $(w\times s) \times (h\times s) \times 64$ & conv & $64\times3\times3\times 64$ & $(w\times s) \times (h\times s) \times 64$ \\
 $(w\times s) \times (h\times s) \times 64$ & conv & $64\times3\times3\times 64$ & $(w\times s) \times (h\times s) \times 64$ \\
 $(w\times s) \times (h\times s) \times 64$ & conv & $64\times3\times3\times 64$ & $(w\times s) \times (h\times s) \times 64$ \\
 $(w\times s) \times (h\times s) \times 64$ & conv & $64\times3\times3\times 64$ & $(w\times s) \times (h\times s) \times 64$ \\
 $(w\times s) \times (h\times s) \times 64$ & conv & $64\times3\times3\times 64$ & $(w\times s) \times (h\times s) \times 64$ \\
 $(w\times s) \times (h\times s) \times 64$ & conv & $64\times3\times3\times 64$ & $(w\times s) \times (h\times s) \times 64$ \\
 $(w\times s)\times (h\times s) \times 64$ & conv & $64\times3\times3\times 25$ & $(w\times s)\times (h\times s) \times 25$ \\
 \hline 
 $(w\times s) \times (h\times s) \times 1 $ & adaptive &~ &~ \\
 $\&$ & image &~ &$(w\times s) \times (h\times s) \times 1$ \\
 $(w\times s) \times (h\times s) \times 25$ & resampling &~ &~ \\
\hline
 $(w\times s) \times (h\times s) \times 1$ & concat &~ & $(w\times s) \times (h\times s) \times 2$ \\
 $(w\times s) \times (h\times s) \times 1$ &~ &~ &~ \\
\hline 
 $(w\times s) \times (h\times s) \times 2$ & conv (shared) & $2\times3\times3\times 64$ & $(w\times s) \times (h\times s) \times 64$ \\
 $(w\times s) \times (h\times s) \times 64$ & conv (shared) & $64\times3\times3\times 64$ & $(w\times s) \times (h\times s) \times 64$ \\
 $(w\times s) \times (h\times s) \times 64$ & conv (shared) & $64\times3\times3\times 64$ & $(w\times s) \times (h\times s) \times 64$ \\
 $(w\times s) \times (h\times s) \times 64$ & conv (shared) & $64\times3\times3\times 64$ & $(w\times s) \times (h\times s) \times 64$ \\
 $(w\times s) \times (h\times s) \times 64$ & conv (shared) & $64\times3\times3\times 64$ & $(w\times s) \times (h\times s) \times 64$ \\
 $(w\times s) \times (h\times s) \times 64$ & conv (shared) & $64\times3\times3\times 64$ & $(w\times s) \times (h\times s) \times 64$ \\
 $(w\times s) \times (h\times s) \times 64$ & conv (shared) & $64\times3\times3\times 64$ & $(w\times s) \times (h\times s) \times 64$ \\
 $(w\times s) \times (h\times s) \times 64$ & conv (shared) & $64\times3\times3\times 64$ & $(w\times s) \times (h\times s) \times 64$ \\
 $(w\times s) \times (h\times s) \times 64$ & conv (shared) & $64\times3\times3\times 64$ & $(w\times s) \times (h\times s) \times 64$ \\
 $(w\times s) \times (h\times s) \times 64$ & conv (shared) & $64\times3\times3\times 64$ & $(w\times s) \times (h\times s) \times 64$ \\
 $(w\times s)\times (h\times s) \times 64$ & conv (shared) & $64\times3\times3\times 25$ & $(w\times s)\times (h\times s) \times 25$ \\
\hline 
 $(w\times s) \times (h\times s) \times 1 $ & adaptive &~ &~ \\
 $\&$ & image &~ &$(w\times s) \times (h\times s) \times 1$ \\
 $(w\times s) \times (h\times s) \times 25$ & resampling &~ &~ \\
\hline
\end{tabular}
}
\end{center}
\label{tab:network_recur2_10}
\end{table}

\section{Visualization of the estimated interpolation kernels for SISR}
\label{sec:vis_dyn_filt}
In this section, we show more examples of estimated interpolation kernels for single image super-resolution in Figure~\ref{fig:dyn_filt_baby}, Figure~\ref{fig:dyn_filt_bird}, Figure~\ref{fig:dyn_filt_head} and Figure~\ref{fig:dyn_filt_woman}.
\begin{figure*}[tp]
\centering
\scalebox{0.9}{
	\begin{tabular}[t]{c@{\hspace{0.005\linewidth}}c@{\hspace{0.005\linewidth}}c}
		\includegraphics[width=0.32\linewidth]{./images/supp/filter_visualization/baby_s3_im.png}
        &
        \includegraphics[width=0.32\linewidth]{./images/supp/filter_visualization/baby_s3_im_nn.png}
        &
        \includegraphics[width=0.32\linewidth]{./images/supp/filter_visualization/baby_s3_out.png}
        \\     
        \footnotesize{(a) HR image}
        &
        \footnotesize{(b) NN resized image}
        &
        \footnotesize{(c) Our result}
        \\          
        \includegraphics[width=0.32\linewidth]{./images/supp/filter_visualization/baby_s3_filter_6.png}
        &
        \includegraphics[width=0.32\linewidth]{./images/supp/filter_visualization/baby_s3_filter_7.png}
        &
        \includegraphics[width=0.32\linewidth]{./images/supp/filter_visualization/baby_s3_filter_8.png}
        \\
        \includegraphics[width=0.32\linewidth]{./images/supp/filter_visualization/baby_s3_filter_11.png}
        &
        \includegraphics[width=0.32\linewidth]{./images/supp/filter_visualization/baby_s3_filter_12.png}
        &
        \includegraphics[width=0.32\linewidth]{./images/supp/filter_visualization/baby_s3_filter_13.png}
        \\
        \includegraphics[width=0.32\linewidth]{./images/supp/filter_visualization/baby_s3_filter_16.png}
        &
        \includegraphics[width=0.32\linewidth]{./images/supp/filter_visualization/baby_s3_filter_17.png}
        &
        \includegraphics[width=0.32\linewidth]{./images/supp/filter_visualization/baby_s3_filter_18.png}
        \\
        ~ & \footnotesize{(d) Estimated filters} & ~
	\end{tabular}	
}    
    \caption{Visualization of the feature maps corresponding to estimated $5\times 5$ interpolation kernels for $3\times$ super-resolution. Note that we only visualize the inner $3\times 3$ part of the $5\times 5$ kernel since most of the outer part is close to zero. It is recommended to see (d) by zooming in in the electronic version. 
    }
    \label{fig:dyn_filt_baby}
\end{figure*}
\begin{figure*}[tp]
\centering
\scalebox{0.9}{
    \begin{tabular}[t]{c@{\hspace{0.005\linewidth}}c@{\hspace{0.005\linewidth}}c}
		\includegraphics[width=0.32\linewidth]{./images/supp/filter_visualization/bird_s3_im.png}
        &
        \includegraphics[width=0.32\linewidth]{./images/supp/filter_visualization/bird_s3_im_nn.png}
        &
        \includegraphics[width=0.32\linewidth]{./images/supp/filter_visualization/bird_s3_out.png}
        \\     
        \footnotesize{(a) HR image}
        &
        \footnotesize{(b) NN resized image}
        &
        \footnotesize{(c) Our result}
        \\          
        \includegraphics[width=0.32\linewidth]{./images/supp/filter_visualization/bird_s3_filter_6.png}
        &
        \includegraphics[width=0.32\linewidth]{./images/supp/filter_visualization/bird_s3_filter_7.png}
        &
        \includegraphics[width=0.32\linewidth]{./images/supp/filter_visualization/bird_s3_filter_8.png}
        \\
        \includegraphics[width=0.32\linewidth]{./images/supp/filter_visualization/bird_s3_filter_11.png}
        &
        \includegraphics[width=0.32\linewidth]{./images/supp/filter_visualization/bird_s3_filter_12.png}
        &
        \includegraphics[width=0.32\linewidth]{./images/supp/filter_visualization/bird_s3_filter_13.png}
        \\
        \includegraphics[width=0.32\linewidth]{./images/supp/filter_visualization/bird_s3_filter_16.png}
        &
        \includegraphics[width=0.32\linewidth]{./images/supp/filter_visualization/bird_s3_filter_17.png}
        &
        \includegraphics[width=0.32\linewidth]{./images/supp/filter_visualization/bird_s3_filter_18.png}
        \\
        ~ & \footnotesize{(d) Estimated filters} & ~
	\end{tabular}
}    
    \caption{Visualization of the feature maps corresponding to estimated $5\times 5$ interpolation kernels for $3\times$ super-resolution. Note that we only visualize the inner $3\times 3$ part of the $5\times 5$ kernel since most of the outer part is close to zero. It is recommended to see (d) by zooming in in the electronic version. 
    }
    \label{fig:dyn_filt_bird}
\end{figure*}
\begin{figure*}[tp]
\centering
\scalebox{0.9}{
    \begin{tabular}[t]{c@{\hspace{0.005\linewidth}}c@{\hspace{0.005\linewidth}}c}
		\includegraphics[width=0.32\linewidth]{./images/supp/filter_visualization/head_s3_im.png}
        &
        \includegraphics[width=0.32\linewidth]{./images/supp/filter_visualization/head_s3_im_nn.png}
        &
        \includegraphics[width=0.32\linewidth]{./images/supp/filter_visualization/head_s3_out.png}
        \\     
        \footnotesize{(a) HR image}
        &
        \footnotesize{(b) NN resized image}
        &
        \footnotesize{(c) Our result}
        \\          
        \includegraphics[width=0.32\linewidth]{./images/supp/filter_visualization/head_s3_filter_6.png}
        &
        \includegraphics[width=0.32\linewidth]{./images/supp/filter_visualization/head_s3_filter_7.png}
        &
        \includegraphics[width=0.32\linewidth]{./images/supp/filter_visualization/head_s3_filter_8.png}
        \\
        \includegraphics[width=0.32\linewidth]{./images/supp/filter_visualization/head_s3_filter_11.png}
        &
        \includegraphics[width=0.32\linewidth]{./images/supp/filter_visualization/head_s3_filter_12.png}
        &
        \includegraphics[width=0.32\linewidth]{./images/supp/filter_visualization/head_s3_filter_13.png}
        \\
        \includegraphics[width=0.32\linewidth]{./images/supp/filter_visualization/head_s3_filter_16.png}
        &
        \includegraphics[width=0.32\linewidth]{./images/supp/filter_visualization/head_s3_filter_17.png}
        &
        \includegraphics[width=0.32\linewidth]{./images/supp/filter_visualization/head_s3_filter_18.png}
        \\
        ~ & \footnotesize{(d) Estimated filters} & ~
	\end{tabular}
}    
    \caption{Visualization of the feature maps corresponding to estimated $5\times 5$ interpolation kernels for $3\times$ super-resolution. Note that we only visualize the inner $3\times 3$ part of the $5\times 5$ kernel since most of the outer part is close to zero. It is recommended to see (d) by zooming in in the electronic version. 
    }
    \label{fig:dyn_filt_head}
\end{figure*}
\begin{figure*}[tp]
\centering
\scalebox{0.65}{
    \begin{tabular}[t]{c@{\hspace{0.005\linewidth}}c@{\hspace{0.005\linewidth}}c}
		\includegraphics[width=0.32\linewidth]{./images/supp/filter_visualization/woman_s3_im.png}
        &
        \includegraphics[width=0.32\linewidth]{./images/supp/filter_visualization/woman_s3_im_nn.png}
        &
        \includegraphics[width=0.32\linewidth]{./images/supp/filter_visualization/woman_s3_out.png}
        \\     
        \footnotesize{(a) HR image}
        &
        \footnotesize{(b) NN resized image}
        &
        \footnotesize{(c) Our result}
        \\          
        \includegraphics[width=0.32\linewidth]{./images/supp/filter_visualization/woman_s3_filter_6.png}
        &
        \includegraphics[width=0.32\linewidth]{./images/supp/filter_visualization/woman_s3_filter_7.png}
        &
        \includegraphics[width=0.32\linewidth]{./images/supp/filter_visualization/woman_s3_filter_8.png}
        \\
        \includegraphics[width=0.32\linewidth]{./images/supp/filter_visualization/woman_s3_filter_11.png}
        &
        \includegraphics[width=0.32\linewidth]{./images/supp/filter_visualization/woman_s3_filter_12.png}
        &
        \includegraphics[width=0.32\linewidth]{./images/supp/filter_visualization/woman_s3_filter_13.png}
        \\
        \includegraphics[width=0.32\linewidth]{./images/supp/filter_visualization/woman_s3_filter_16.png}
        &
        \includegraphics[width=0.32\linewidth]{./images/supp/filter_visualization/woman_s3_filter_17.png}
        &
        \includegraphics[width=0.32\linewidth]{./images/supp/filter_visualization/woman_s3_filter_18.png}
        \\
        ~ & \footnotesize{(d) Estimated filters} & ~
	\end{tabular}
}    
    \caption{Visualization of the feature maps corresponding to estimated $5\times 5$ interpolation kernels for $3\times$ super-resolution. Note that we only visualize the inner $3\times 3$ part of the $5\times 5$ kernel since most of the outer part is close to zero. It is recommended to see (d) by zooming in in the electronic version. 
    }
    \label{fig:dyn_filt_woman}
\end{figure*}

%-------------------------------------------------------------------------
\section{Visualization of the estimated interpolation kernels for joint image filtering}
\label{sec:vis_dyn_filt_joint}
In this section, we also show some examples of estimated interpolation kernels for depth image super-resolution with guidance in Figure~\ref{fig:dyn_filt_1}, Figure~\ref{fig:dyn_filt_5}, Figure~\ref{fig:dyn_filt_90} and Figure~\ref{fig:dyn_filt_156}. We can see that the feature map the middle still has higher values than the others. However, we find that the feature maps next to the middle one do not show vertical and horizontal strips, which is different from the interpolation kernels estimated for single image super-resolution.
\begin{figure*}[t]
\centering
\begin{subfigure}[t]
	\centering
    \begin{tabular}{c@{\hspace{0.005\linewidth}}c@{\hspace{0.005\linewidth}}c@{\hspace{0.005\linewidth}}c}
		\includegraphics[width=0.24\linewidth]{./images/supp/joint_filter_visualization/1/1_s8_gt.png}
        &
        \includegraphics[width=0.24\linewidth]{./images/supp/joint_filter_visualization/1/1_s8_guidance.png}
        &
        \includegraphics[width=0.24\linewidth]{./images/supp/joint_filter_visualization/1/1_s8_in_nn.png}
        &
        \includegraphics[width=0.24\linewidth]{./images/supp/joint_filter_visualization/1/1_s8_output.png}
        \\     
        \footnotesize{(a) HR image}
        &
        \footnotesize{(b) Guidance}
        &
        \footnotesize{(c) NN resized image}
        &
        \footnotesize{(d) Our result}
    \end{tabular}  
\end{subfigure}    
    \begin{tabular}{c@{\hspace{0.005\linewidth}}c@{\hspace{0.005\linewidth}}c}
        \\
        \includegraphics[width=0.32\linewidth]{./images/supp/joint_filter_visualization/1/1_s8_filter_6.png}
        &
        \includegraphics[width=0.32\linewidth]{./images/supp/joint_filter_visualization/1/1_s8_filter_7.png}
        &
        \includegraphics[width=0.32\linewidth]{./images/supp/joint_filter_visualization/1/1_s8_filter_8.png}
        \\
        \includegraphics[width=0.32\linewidth]{./images/supp/joint_filter_visualization/1/1_s8_filter_11.png}
        &
        \includegraphics[width=0.32\linewidth]{./images/supp/joint_filter_visualization/1/1_s8_filter_12.png}
        &
        \includegraphics[width=0.32\linewidth]{./images/supp/joint_filter_visualization/1/1_s8_filter_13.png}
        \\
        \includegraphics[width=0.32\linewidth]{./images/supp/joint_filter_visualization/1/1_s8_filter_16.png}
        &
        \includegraphics[width=0.32\linewidth]{./images/supp/joint_filter_visualization/1/1_s8_filter_17.png}
        &
        \includegraphics[width=0.32\linewidth]{./images/supp/joint_filter_visualization/1/1_s8_filter_18.png}
        \\
        ~ & \footnotesize{(e) Estimated filters} & ~
	\end{tabular}
    \caption{Visualization of the feature maps corresponding to estimated $5\times 5$ interpolation kernels for $8\times$ super-resolution. Note that we only visualize the inner $3\times 3$ part of the $5\times 5$ kernel since most of the outer part is close to zero. It is recommended to see (e) by zooming in in the electronic version. 
    }
    \label{fig:dyn_filt_1}
\end{figure*}
\begin{figure*}[t]
\centering
\begin{subfigure}[t]
	\centering
    \begin{tabular}{c@{\hspace{0.005\linewidth}}c@{\hspace{0.005\linewidth}}c@{\hspace{0.005\linewidth}}c}
		\includegraphics[width=0.24\linewidth]{./images/supp/joint_filter_visualization/5/5_s8_gt.png}
        &
        \includegraphics[width=0.24\linewidth]{./images/supp/joint_filter_visualization/5/5_s8_guidance.png}
        &
        \includegraphics[width=0.24\linewidth]{./images/supp/joint_filter_visualization/5/5_s8_in_nn.png}
        &
        \includegraphics[width=0.24\linewidth]{./images/supp/joint_filter_visualization/5/5_s8_output.png}
        \\     
        \footnotesize{(a) HR image}
        &
        \footnotesize{(b) Guidance}
        &
        \footnotesize{(c) NN}
        &
        \footnotesize{(d) Our result}
    \end{tabular}  
\end{subfigure}    
    \begin{tabular}{c@{\hspace{0.005\linewidth}}c@{\hspace{0.005\linewidth}}c}
        \\
        \includegraphics[width=0.32\linewidth]{./images/supp/joint_filter_visualization/5/5_s8_filter_6.png}
        &
        \includegraphics[width=0.32\linewidth]{./images/supp/joint_filter_visualization/5/5_s8_filter_7.png}
        &
        \includegraphics[width=0.32\linewidth]{./images/supp/joint_filter_visualization/5/5_s8_filter_8.png}
        \\
        \includegraphics[width=0.32\linewidth]{./images/supp/joint_filter_visualization/5/5_s8_filter_11.png}
        &
        \includegraphics[width=0.32\linewidth]{./images/supp/joint_filter_visualization/5/5_s8_filter_12.png}
        &
        \includegraphics[width=0.32\linewidth]{./images/supp/joint_filter_visualization/5/5_s8_filter_13.png}
        \\
        \includegraphics[width=0.32\linewidth]{./images/supp/joint_filter_visualization/5/5_s8_filter_16.png}
        &
        \includegraphics[width=0.32\linewidth]{./images/supp/joint_filter_visualization/5/5_s8_filter_17.png}
        &
        \includegraphics[width=0.32\linewidth]{./images/supp/joint_filter_visualization/5/5_s8_filter_18.png}
        \\
        ~ & \footnotesize{(e) Estimated filters} & ~
	\end{tabular}
    \caption{Visualization of the feature maps corresponding to estimated $5\times 5$ interpolation kernels for $8\times$ super-resolution. Note that we only visualize the inner $3\times 3$ part of the $5\times 5$ kernel since most of the outer part is close to zero. It is recommended to see (e) by zooming in in the electronic version. 
    }
    \label{fig:dyn_filt_5}
\end{figure*}
\begin{figure*}[t]
\centering
\begin{subfigure}[t]
	\centering
    \begin{tabular}{c@{\hspace{0.005\linewidth}}c@{\hspace{0.005\linewidth}}c@{\hspace{0.005\linewidth}}c}
		\includegraphics[width=0.24\linewidth]{./images/supp/joint_filter_visualization/90/90_s8_gt.png}
        &
        \includegraphics[width=0.24\linewidth]{./images/supp/joint_filter_visualization/90/90_s8_guidance.png}
        &
        \includegraphics[width=0.24\linewidth]{./images/supp/joint_filter_visualization/90/90_s8_in_nn.png}
        &
        \includegraphics[width=0.24\linewidth]{./images/supp/joint_filter_visualization/90/90_s8_output.png}
        \\     
        \footnotesize{(a) HR image}
        &
        \footnotesize{(b) Guidance}
        &
        \footnotesize{(c) NN}
        &
        \footnotesize{(d) Our result}
    \end{tabular}  
\end{subfigure}    
    \begin{tabular}{c@{\hspace{0.005\linewidth}}c@{\hspace{0.005\linewidth}}c}
        \\
        \includegraphics[width=0.32\linewidth]{./images/supp/joint_filter_visualization/90/90_s8_filter_6.png}
        &
        \includegraphics[width=0.32\linewidth]{./images/supp/joint_filter_visualization/90/90_s8_filter_7.png}
        &
        \includegraphics[width=0.32\linewidth]{./images/supp/joint_filter_visualization/90/90_s8_filter_8.png}
        \\
        \includegraphics[width=0.32\linewidth]{./images/supp/joint_filter_visualization/90/90_s8_filter_11.png}
        &
        \includegraphics[width=0.32\linewidth]{./images/supp/joint_filter_visualization/90/90_s8_filter_12.png}
        &
        \includegraphics[width=0.32\linewidth]{./images/supp/joint_filter_visualization/90/90_s8_filter_13.png}
        \\
        \includegraphics[width=0.32\linewidth]{./images/supp/joint_filter_visualization/90/90_s8_filter_16.png}
        &
        \includegraphics[width=0.32\linewidth]{./images/supp/joint_filter_visualization/90/90_s8_filter_17.png}
        &
        \includegraphics[width=0.32\linewidth]{./images/supp/joint_filter_visualization/90/90_s8_filter_18.png}
        \\
        ~ & \footnotesize{(e) Estimated filters} & ~
	\end{tabular}
    \caption{Visualization of the feature maps corresponding to estimated $5\times 5$ interpolation kernels for $8\times$ super-resolution. Note that we only visualize the inner $3\times 3$ part of the $5\times 5$ kernel since most of the outer part is close to zero. It is recommended to see (e) by zooming in in the electronic version. 
    }
    \label{fig:dyn_filt_90}
\end{figure*}
\begin{figure*}[t]
\centering
\begin{subfigure}[t]
	\centering
    \begin{tabular}{c@{\hspace{0.005\linewidth}}c@{\hspace{0.005\linewidth}}c@{\hspace{0.005\linewidth}}c}
		\includegraphics[width=0.24\linewidth]{./images/supp/joint_filter_visualization/156/156_s8_gt.png}
        &
        \includegraphics[width=0.24\linewidth]{./images/supp/joint_filter_visualization/156/156_s8_guidance.png}
        &
        \includegraphics[width=0.24\linewidth]{./images/supp/joint_filter_visualization/156/156_s8_in_nn.png}
        &
        \includegraphics[width=0.24\linewidth]{./images/supp/joint_filter_visualization/156/156_s8_output.png}
        \\     
        \footnotesize{(a) HR image}
        &
        \footnotesize{(b) Guidance}
        &
        \footnotesize{(c) NN}
        &
        \footnotesize{(d) Our result}
    \end{tabular}
\end{subfigure}    
    \begin{tabular}{c@{\hspace{0.005\linewidth}}c@{\hspace{0.005\linewidth}}c}
        \\
        \includegraphics[width=0.32\linewidth]{./images/supp/joint_filter_visualization/156/156_s8_filter_6.png}
        &
        \includegraphics[width=0.32\linewidth]{./images/supp/joint_filter_visualization/156/156_s8_filter_7.png}
        &
        \includegraphics[width=0.32\linewidth]{./images/supp/joint_filter_visualization/156/156_s8_filter_8.png}
        \\
        \includegraphics[width=0.32\linewidth]{./images/supp/joint_filter_visualization/156/156_s8_filter_11.png}
        &
        \includegraphics[width=0.32\linewidth]{./images/supp/joint_filter_visualization/156/156_s8_filter_12.png}
        &
        \includegraphics[width=0.32\linewidth]{./images/supp/joint_filter_visualization/156/156_s8_filter_13.png}
        \\
        \includegraphics[width=0.32\linewidth]{./images/supp/joint_filter_visualization/156/156_s8_filter_16.png}
        &
        \includegraphics[width=0.32\linewidth]{./images/supp/joint_filter_visualization/156/156_s8_filter_17.png}
        &
        \includegraphics[width=0.32\linewidth]{./images/supp/joint_filter_visualization/156/156_s8_filter_18.png}
        \\
        ~ & \footnotesize{(e) Estimated filters} & ~
	\end{tabular}
    \caption{Visualization of the feature maps corresponding to estimated $5\times 5$ interpolation kernels for $8\times$ super-resolution. Note that we only visualize the inner $3\times 3$ part of the $5\times 5$ kernel since most of the outer part is close to zero. It is recommended to see (e) by zooming in in the electronic version. 
    }
    \label{fig:dyn_filt_156}
\end{figure*}

{\small
\bibliographystyle{ieee}
\bibliography{egbib}
}